\documentclass[healthcaree,review,submit,moreauthors,pdftex,outdir=./]{Definitions/mdpi}

\usepackage{tabulary,xcolor}
\usepackage{tcolorbox}
\usepackage{enumitem}
\usepackage{amsfonts,amsmath,amssymb}
\usepackage[T1]{fontenc}
\usepackage{longtable}
\usepackage{amsfonts,amsmath,amssymb}
\firstpage{1} 
\pubvolume{1}
\issuenum{1}
\articlenumber{0}
\pubyear{2021}
\copyrightyear{2020}
\datereceived{} 
\dateaccepted{} 
\datepublished{} 
\hreflink{https://doi.org/} 

\Title{A Comprehensive Survey of Image-Based  Food Recognition and Volume Estimation Methods for Dietary Assessment}

\TitleCitation{Title}

\Author{Ghalib Tahir $^{1,\dagger,\ddagger}$ and Chu Kiong Loo $^{1,\ddagger}$}

\AuthorNames{Firstname Lastname, Firstname Lastname and Firstname Lastname}

\AuthorCitation{Lastname, F.; Lastname, F.; Lastname, F.}

\address{%
$^{1}$ \quad Affiliation 1; 12mscsgtahir@seecs.edu.pk,ghalib@siswa.um.edu.my\\
$^{2}$ \quad Affiliation 2; ckloo.um@um.edu.my}

\firstnote{Current address: Department of Artificial Intelligence, Faculty of Computer Science and Information Technology\unskip, University of  Malaya, 50603 Kuala Lumpur, Malaysia} 

\abstract{ Dietary studies showed that dietary-related problem such as obesity is associated with other chronic diseases like hypertension, irregular blood sugar levels, and increased risk of heart attacks. The primary cause of these problems is poor lifestyle choices and unhealthy dietary habits, which are manageable using interactive mHealth apps. However, traditional dietary monitoring systems using manual food logging suffer from imprecision, underreporting, time consumption, and low adherence. Recent dietary monitoring systems tackle these challenges by automatic assessment of dietary intake through machine learning methods.  This survey discusses the most performing methodologies that have been developed so far for automatic food recognition and volume estimation. First, we will present the rationale of visual-based methods for food recognition. The core of the paper is the presentation, discussion and evaluation of these methods on popular food image databases. Following that, we discussed the mobile applications that are implementing these methods. The survey ends with a discussion of research gaps and open issues in this area.}


\begin{document}

\nolinenumbers

\section{Introduction}

Despite recent advancements in medicine, the number of people affected by chronic diseases is still significantly large. This rate is primarily due to their unhealthy lifestyles and irregular eating patterns. Some of the more notable chronic diseases include obesity, hypertension, blood sugar, cardiovascular diseases, and different kinds of cancers. Out of these, obesity and weight issues are becoming increasingly common around the globe.  It affects almost every part of the world, from middle to low-income countries. In 2016, 1.9 billion adults 18 years and older were overweight, while 650 million were obese. With time, children are also becoming affected by obesity at an alarming rate. According to World Health Organization (WHO), over 340 million children and adolescents between 5 and 19 years were overweight or obese. \cite{intro-1}.  
\par The prevalence of these alarming statistics poses a serious concern. However, determining the effective remedial measures depends on different factors, ranging from a person’s genetics to lifestyle choices. To cope with chronic weight problems, people usually keep notes to track their dietary intake. In turn, dieticians require these records to estimate a patient’s nutrient consumption. Many dietary mobile applications automate this process by developing food recognition and volume estimation models that directly classify food categories, ingredients, estimate nutrients from smartphone camera pictures. 
\par However, automatic food recognition using a smartphone camera in the real world is considered a multi-dimensional problem. Unlike other image classification problems, food recognition is a complex task that involves several challenges. There is no spatial layout information that it can exploits like, in the case of the human body, there is a spatial relationship between body parts. The head is always present over the trunk of the human body \cite{intro-2,intro-3,intro-4} and feet towards the lower end. Similarly, the non-rigid structure of the food and intra-source variations make it even more complicated to classify food items correctly as preparation methods and cooking styles vary from region to region.  Moreover, inter-class ambiguity is also a source of potential recognition problems as different food items may look very similar (e.g. soups). Moreover, in many dishes, some ingredients are concealed from view that can limit the performance of food ingredient classification models.
\par In addition to this, image quality from the smartphone camera is dependent on different types of cameras, lighting conditions, and orientations. As a result, the poor performance of food recognition models is highly susceptible to image distortions. 
\par Despite these challenges, many food images possess distinctive properties to distinguish one food type from another. Firstly, the visual representations of food images are of fundamental importance as it significantly impacts classification performance. Therefore, many food recognition methods employ handcrafted features such as shape, colour, texture, local. Recent techniques are using deep visual features for image representation. Some of these methods implement a combination of handcrafted and deep visual features for image feature representation.  Secondly, for enhanced classification performance and reduced computational complexity, an appropriate selection of attributes is essential for removing redundant features from feature vectors. Finally, wisely selecting classification techniques is crucial to address food recognition challenges effectively. 
\par Similarly, manual logging of food volume is a tedious task and involves a high rate of human error by as much as 30\% \cite{intro-v1,intro-v2,intro-v3,intro-v4,intro-v5, intro-v6}. Several solutions are proposed whose aim is to estimate food volume from smartphone camera pictures. Previous studies \cite{intro-v7} shows that using a mobile phone camera for food volume estimation increases the accuracy of calories estimation. Some methods involve capturing a single image, while multiple views are needed to determine accurate volume in other techniques. The food volume estimation process involves the following two steps 1) multiple images or a single image from a mobile camera is needed 2) computation of food volume from 3d construction or calibration object. Regardless of other volume estimation tasks, food volume estimation is a complex task with many specific challenges. Many foods have variations in shape and appearance due to shape and eating conditions.

\par
\begin{figure}[!htbp]
\centering
\includegraphics[width=4.3in]{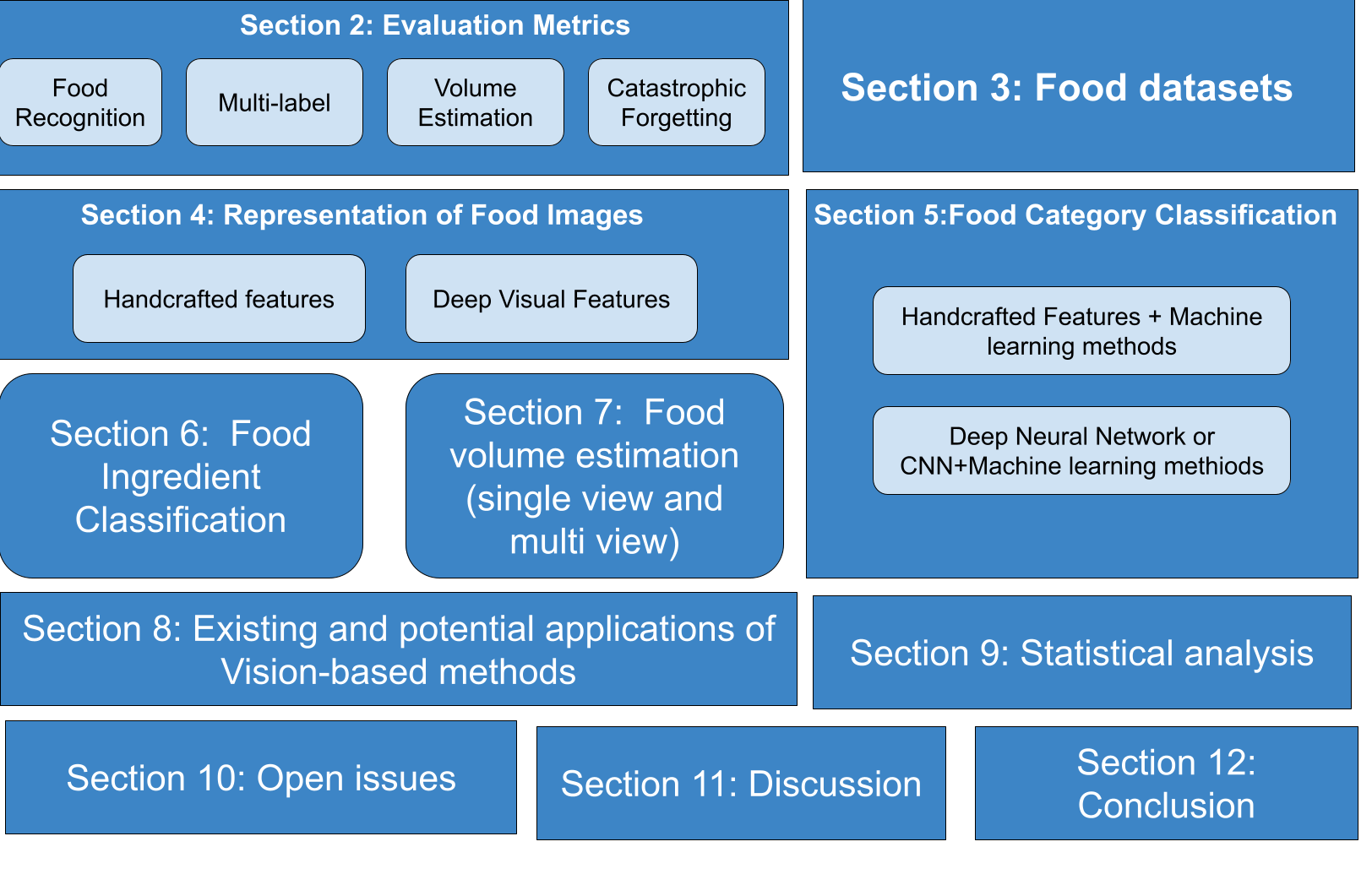}
\caption{Scope and taxonomy of this survey paper}
\label{taxonomy}
\end{figure}

 The following research paper aims to scrutinize state-of-the-art vision-based approaches for dietary assessment. Figure \ref{taxonomy}. represents the detailed scope and taxonomy of our survey study. The contribution of this survey is summarized as follows:
\par 1) We have briefly explored existing food databases for evaluating vision-based approaches and performance measures to thoroughly investigate food recognition, ingredient detection and volume estimation methods.
\par 2) We present an extensive review of food recognition techniques, including traditional methods with handcrafted features and modern deep learning-based approaches.
\par 3) We provided deep insight into multi-label methods for food ingredient classification.
\par 4) We surveyed most performing single view and multiview methods for food volume estimation.
\par 5) We presented existing mobile applications which implements these approaches and other potential applications of vision-based methods in health care.
\par 6) We analyzed open issues and suggested possible solutions to overcome the limitations of the existing methodologies.
\par The rest of the article is organized as follows.  Section II and III examine evaluation metrics and existing datasets. Section IV examines feature extraction methods for food image representation including, handcrafted and deep visual features. In sections V and VI, we presented the most performing classifiers for food categorization and ingredient detection. Section VII represents the food volume estimation methods.  In section VIII,  we provide brief information about mobile applications implementing these methods and other potential applications.  Section IX and X summarizes statistical analysis and open issues. Conclusively, we highlight our findings and future works related to this topic.
\section{Evaluation Metrics}
\subsection{Evaluation Metrics for Food Categorization}
\begin{figure}[!htbp]
\centering
\includegraphics[width=3in]{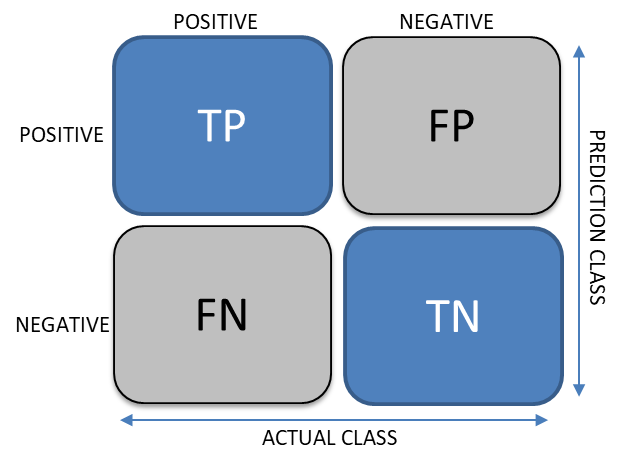}
\caption{Confusion Matrix}
\label{confusion matrix}
\end{figure}

Performance of automatic food recognition models is highly dependent on correct mapping of food images into their respective categories. Therefore, evaluation metrics plays an essential role to determine the correctness of food recognition models. Several metrics have been discussed in literature and their appropriate selection depends on the requirements of specific applications. It has also been observed that a classifier may perform well under one metric but poorly under another metric.  However, intrinsic metrics generally used for better comparisons are ‘Accuracy’, ‘Precision’, ‘Recall’ and ‘F1’ are discussed in detail below.
\\\textbf{Accuracy}
\\Accuracy of a model determines whether a model is being able to predict food classes correctly or how well a certain model can perform generally. Eq. (\ref{eqn_accuracy}) represents mathematical form of accuracy. However, accuracy cannot be used as major performance metric, as it does not serve the purpose when there’s imbalance dataset. Therefore, we have incorporated ‘Precision’, ‘Recall’ and ‘F1 score’ to provide better insights of the results. 
\begin{equation}
\label{eqn_accuracy}
Accuracy\;=\frac{\;(TP+FN)}{(TP+FP+FN+TN)}\times100\;
\end{equation}
\\\textbf{Precision}
\\Precision score can be defined as how often a certain model can correctly predict classified positive values. In simpler words, out of all predicted positive food classes what percentage is truly positive. This score is beneficial when the cost of false positives is high. It is calculated by Eq. (\ref{eqn_precision}).
\begin{equation}
\label{eqn_precision}
Precision\;Score\;=\frac{TP}{(TP\;+FP)}\;\;
\end{equation}
\\\textbf{Recall}
\\Recall score identifies model’s ability to correctly classify food classes. It determines out of total positive food classes what percentage are predicted positives. It provides better insight when cost of false negatives is high. It is computed by using Eq. (\ref{eqn_recall}).
\begin{equation}
\label{eqn_recall}
Recall\;=\;\frac{TP}{(TP+FN)}
\end{equation}
\\\textbf{F1 Score}
\\F1 score represents the harmonic mean of recall and precision's score. It considers both false positives and false negatives, therefore, it performs great on imbalanced datasets. It is calculated by following Eq. (\ref{eqn_f1score}).
\begin{equation}
\label{eqn_f1score}
F1\;Score\;=\;\frac{(2\ast(Precision\ast Recall))}{Precision+Recall}\;
\end{equation}
\\\textbf{Confusion Matrix:}
\\Confusion Matrix is a widely used approach to summarize the performance of a classification model in machine learning. In some cases, classification accuracy alone can be misleading especially when there are more than two classes in a dataset or if there are unequal number of observations present in classes. Therefore, confusion matrix provides clear picture of actual and predicted classes obtained by classification model. Confusion matrix is basically a two dimensional matrix, where each row represents example of an actual class and each column represents state of predicted class. TP stands for true positive, TN represents the number of true negative, FP is the number of false positive and FN represents false negative in confusion matrix shown in Figure(\ref{confusion matrix}) .

\subsection{Catastrophic Forgetting During Progressive Learning}
Catastrophic forgetting during open-ended learning measures the algorithm's ability to add new neurons or classes corresponding to novel incoming concepts. Kemker et al. \cite{catas_1} and Chaudry et al. \cite{catas_2} proposed five measures of catastrophic forgetting to achieve this objective.
\paragraph{Intransigence}
It refers to the difference of classification performance between the reference model trained by batch learning technique and the model trained on feature vectors using incremental learning protocol.  The negative intransigence represents that incrementally learning a set of classes is improving performance.  Eq. (\ref{eqn_catas0}) denotes its mathematical form.
\begin{equation}
 \label{eqn_catas0}
    l_k= a_k^*- a_k, _k
\end{equation}

\paragraph{Forgetting}

It refers to the difference between the maximum classification performance of a particular session in previous sessions and its classification performance in the current session. Eq. (\ref{eqn_catas000}) computes the average forgetting of the network up to the $ k^{th} \;$ session.
\begin{equation}
     \label{eqn_catas000}
    \begin{array}{l}f_j^k=max_{1\in\{1,\dots\dots,K-1\}}\;a_{i,j}-a_{(k,j)},j>k\\\\      F_k=\frac1{k-1}\sum\nolimits_{j=1}^{k-1}f_j^k\;\end{array}
\end{equation}
\paragraph{Base Session}
It refers to the classification performance on test images of base classes in the current session as shown by eq. (\ref{eqn_catas1})
\begin{equation}
     \label{eqn_catas1}
    \Omega_{base}=\frac{1\;}{k-1}\sum\nolimits_{j=2}^k\frac{a_{j,1}}{a_{ideal}}\;
\end{equation}
\paragraph{New Session}
It is the ability of a model to recall newly learning knowledge as shown in the eq. (\ref{eqn_catas2}).
\begin{equation}
 \label{eqn_catas2}
    \Omega_{new}=\frac{1\;}{k-1}\sum\nolimits_{j=2}^ka_{j,j}\;
\end{equation}
\paragraph{All Session}
It refers to the retention of the previously learned information by the network when learning new classes as computed by eq. (\ref{eqn_catas3}).
\begin{equation}
 \label{eqn_catas3}
    \Omega_{all}=\frac{1\;}{k-1}\sum\nolimits_{j=2}^k\frac{a_{j,all}}{a_{ideal}}\;
\end{equation}
\subsection{Evaluation Metrics for Food Ingredient Classification}
Similarly, food ingredient recognition is equally important for dietary assessment applications.As food categorization is limited to the classification of generic food items present in the food images; food ingredient recognition and classification provides deep insights into the caloric content present in the food image. Therefore, food ingredient recognition applications widely incorporate multi-label classification. \cite{ml2}. Since, food ingredient recognition is considered as multi-label problem as food image usually contains more than one ingredient. Therefore, evaluation metrics generally used for multi-label classification are different from traditional single-label classification. Following are the performance metrics are used by food ingredient recognition models.
\\Consider $ {x_i,Y_i} \;$ with L number of labels as training datasets. Let assume MLC is the training method and $ Z_i = MLC(x_i) \;$  is the output labels (Ingredients) predicted by the classification method.
\\\textbf{Precision}
\\Precision is the ratio of correctly predicted labels to total number of actual labels, averaged across all instances. Eq(\ref{eqn_precision_ml}) represents precision for food ingridient classification.
\begin{equation}
\label{eqn_precision_ml}
Precision\;=\;\frac1N\sum_{i=1}^N\;\left(\frac{MLC(x_i)\;\cap Y_i}{MLC(x_i)}\right)
\end{equation}
\\\textbf{Recall}
\\Recall is computed by Eq(\ref{eqn_recall_ml}). It is the ratio of correctly predicted labels to the total number of predicted labels. 
\begin{equation}
\label{eqn_recall_ml}
Recall\;=\;\frac1N\sum_{i=1}^N\;\left(\frac{MLC(x_i)\;\cap Y_i}{MLC(Y_i)}\right)
\end{equation}
\\\textbf{F1 Score}
\\Finally, F1 score is the harmonic mean of the precision and recall. Eq(\ref{eqn_f1_ml}) represents the F1 score.
\begin{equation}
\label{eqn_f1_ml}
F1\;Score\;=\;\frac1N\sum_{i=1}^N\;\left(\frac{2\ast\;\left|MLC(x_i)\;\cap Y_i\right|}{\left|MLC(x_i)\right|\;+\;\left|Y_i\right|}\right)
\end{equation}
\subsection{Evaluation Metrics for Food Volume Estimation}
Similarly, various studies related to food volume estimation use ground truth values to compare the accuracy of their proposed methods to determine the accurate food volume \cite{pm1} \cite{pm2} \cite{pm3} \cite{pm4} \cite{pm5} \cite{pm6} \cite{pm7} \cite{pm8} \cite{pm9} \cite{pm10} \cite{pm11} \cite{pm12}. Unfortunately, there is no dataset available till date for accurate measurement of food volume. Nevertheless, method proposed by \cite{pm13} uses controlled experiments that require participants to click images before and after their meal to compute consumed calories, which are later compared with ground truth values. Similarly, \cite{pm14} has incorporated different food models to determine the true volume, however, various models failed to provide accurate information. Therefore, they implemented water displacement method which requires mean of three readings to find out true volume.
Also, most of the studies used following equations to compute the relative error and estimate the accuracy of the method
\begin{equation}
    e\;=\;\left|v\;-\;v_{approx}\right|
\end{equation}
Where v is the actual volume and $v_{approx}$ is the approximate volume
\begin{equation}
    e\;=\;\frac1N{\textstyle\sum_{i=1}^n}\frac{\left|w_{i\;-}\;w_g\right|}{w_g}
\end{equation}
Where N is the number of food items; $w_i$  is the estimated weight of the food item;$w_g$  is the ground truth value of the food.

\section{Datasets Used for Food Recognition}
\par Performance of feature extraction and classification techniques is highly dependent on the detail-oriented collection of images, which,  in our case, happen to be food images. As consolidated large food image datasets, for example, UECFOOD-100, Food-101, UECFOOD-256, UNCIT-FD1200, UNCIT-FD889 are eventually used as benchmarks to collate recognition performance of existing approaches with new classifiers. Such datasets can be distinctive in terms of characteristics, such as the total number of images in a particular dataset, cuisine type, and included food categories. Table \ref{datasets} summarizes the characteristics of food datasets surveyed in this research endeavour.
\par For instance, UECFOOD-100 contains 100 different sorts of food categories, and each food category has a bounding box that indicates the location of the food item in the photograph. Food categories in this dataset mainly belong to popular foods in Japan \cite{dataset2}. Similarly, UECFOOD-256 is another variant of UECFOOD-100.  However, it differs in terms of the number of images as it contains 256 food images of different kinds  \cite{dataset2}. Food-101 contains 101000 real-world images that are classified into 101 food categories. It includes diverse yet visually similar food classes \cite{hf11}. Similarly, the PFID food dataset is composed of 1098 food images from 61 different categories. The PFID collection currently has three instances of 101 fast foods \cite{dataset26}. UNCIT-FD1200 is composed of 4754 food images of 1200 types of dishes captured from actual meals. Each food plate is acquired multiple times, and the overall dataset presents both geometric and photometric variability. Similarly, UNICT-FD 889 dataset has 3583 images \cite{dataset13} of 889 different real food plates captured using mobile devices in uncontrolled scenarios (e.g., different backgrounds and light environmental conditions). Moreover, they capture each dish image in UNICT-FD899 multiple times to ensure geometric and photometric variability (rotation, scale, point of view changes) 
\cite{dataset24}. 
\par Several datasets mainly consist of various food images collected through various sources such as web crawlers, social media platforms like Instagram, Flickr, and Facebook. Also, most of these datasets contain images of foods that are specific to certain regions, such as Vireo-Food 172 \cite{hclassification1} and ChineseFoodNet \cite{hclassification3}. Both datasets contain Chinese dishes. Similarly, Food-50 \cite{hf1}, Food-85 \cite{hf1}, Food log [103], UECFOOD-100 \cite{dataset2} and UECFOOD-256 \cite{hf11} contains Japanese Foods items. Turkish foods-15 \cite{dataset16} is limited to Turkish food items only. Also, the Pakistani Food Dataset \cite{d42} accommodates Pakistani dishes, and the Indian Food Database incorporates Indian cuisines. In addition to this, few datasets only include fruits and vegetables like VegFru \cite{dataset7}, Fruits 360 Dataset \cite{dataset6}, and FruitVeg-81 \cite{ma6}. Furthermore, Table \ref{datasets} provides brief description about food image datasets.  Figure \ref{fig_sampleimages}  shows the sample images from the food datasets.



\begin{table*}[!ht]

\centering
\caption{Food image datasets}
\label{datasets}
\resizebox{\textwidth}{!}{%
\begin{tabular}{llllllllll}
\hline
Authors &
 Year &
  Dataset &
  Food Category &
  \begin{tabular}[c]{@{}l@{}}Total \# images\\/ Class\end{tabular} &
  Image Source  \\ \hline
  S. Godwin et al. \cite{datasetfv4}&
   2006
 &
  Wedge Shape foods dataset & American Foods 
   &
   3 categories
   &
   Controlled environment
   \\ \hline
   
Chen et al. \cite{dataset26}&   2009 &
  PFID &
   American Fast Foods &
  1038(61) &
\begin{tabular}[c]{@{}l@{}}  Fast food data captured in\\ multiple restaurants\end{tabular} &

   \\ \hline
  Mariappan et al. \cite{dataset1} &  2009 &
  TADA &
\begin{tabular}[c]{@{}l@{}}  Artificial And\\ Generic Food \end{tabular} &
  256(11) &
  Controlled environment &

   \\ \hline
Yanai et al. \cite{hf1}  &2010 &
  Food-50 &
  Japanese Foods &
  5000(50) &
  Crawled from web &

\\ \hline
Hoashi et al. \cite{hf1} & 2010 &
  Food-85 &
  Japanese Foods &
  8500(85) &
  Existing food databases &

   \\ \hline
  Miyazaki et al. \cite{pm2} &2011 &
  Foodlog &
  Japanese Foods &
  6512(2000) &
  Captured by users &
  
     \\ \hline
   Marc Bosch et al. \cite{datasetfv1}&
   2011
 &
  FNDDS & American Foods
   &
    7000
   &
   Images of food accquired by users
   \\ \hline
   


  
  Matsuda et al. \cite{dataset2} &2012 &
  UECFOOD-100 &
  Japanese Foods &
  14,361(100) &
  Captured by mobile camera &
  
   \\ \hline
  Chen et al.\cite{hclassification3} &2012 &
  ChineseFoodNet &
  Chinese   dishes. &
  192,000(208) &
  Gathered from web &

   \\ \hline
   M.-Y. Chen et al.  \cite{dataset25}&
  2012
 &
  Chen & Chinese Foods
   &
  5000/50 & \begin{tabular}[c]{@{}l@{}}Crawled from the Internet\end{tabular}
   
   &
  
   \\ \hline
  Bossard et al. \cite{d17}&
  2014 &
  Food-101 &
  American Foods &
  101000(101) &
  Crawled from web 
   \\ \hline
   L. Bossard et al. \cite{dataset10}&
  2014
 &
  ETHZ Food-101 & American Foods
   &
  100,000(101) &
  Crawled from web
   \\ \hline
   Kawano et al. \cite{hf11} &2014 &
  UECFOOD-256 &
  Japanese Foods &
  25,088(256) &
  Captured by mobile camera &

   \\ \hline
    T. Stutz et al. \cite{datasetfv5}&
   2014
 &
 Rice dataset & Generic (Rice)
   &
   1 food type
   &
  Acquired from user
   \\ \hline
Farinella et al. \cite{dataset24}&
  2014
 &
  UNCIT-FD889 & Italian Foods
   & 
  3583(899) & Acquired with a smartphone

   \\ \hline
Meyers et al. \cite{ma4}& 2015  & \begin{tabular}[c]{@{}l@{}}FOOD201-\\Segmented\end{tabular} & American Foods                         & 12625   & Manually annotated dataset   \\ \hline

 Xin Wang et al. \cite{dataset9}
 &2015 &
  UPMC Food-101 & Generic
   &
  100,000(101) &
  Crawled from web 
   
   \\ \hline

    Cioccoa et al. \cite{dataset11} & 
  2015
 &
  UNIMB 2015 & Generic
   &
  2000(15) & \begin{tabular}[c]{@{}l@{}}Using a Samsung Galaxy\\ S3 smartphone\end{tabular}

   \\ \hline
   Shaobo Fang et al. \cite{datasetfv3}&
   2015
 &
  TADA(19 foods) & American Foods
   &
    19 categories
   &
   Controlled environment
   \\ \hline
    
   Xu et al. \cite{dataset4}&
  2015&
  Dishes &
 \begin{tabular}[c]{@{}l@{}} Chinese Restaurant Foods \end{tabular}&
  117,504(3,832) &
  Download from dianping 

   \\ \hline
  Beijbom et al. \cite{hclassification4}& 2015 & Menu-Match        & \begin{tabular}[c]{@{}l@{}}Generic Restaurant Food\end{tabular} & 646(41) & Captured from  social media \\ \hline
Zhou et al. \cite{dataset17} &
  2016 
 &
  Food-975 & Chinese Foods
   &
  37,785(975) & Collected from restaurants

   \\ \hline
J. chen et al. \cite{hclassification1} &2016 &
  Vireo-Food 172 &
  Chinese Foods &
  110,241(172) &
  Downloaded from web 
  \\ \hline
 
Cioccoa et al. \cite{dataset12}&
  2016  &
  UNIMB 2016 &
  Italian Foods &
  1,027(73) &
 \begin{tabular}[c]{@{}l@{}}  Captured from\\ dinning tables\end{tabular}  
  \\ \hline
  Hui Wu et al. \cite{dataset18}&
  2016
 &
  Food500 &
  Generic
   &
  148,408 (508) &Crawled from web
    
   \\ \hline
   Singla et al. \cite{dataset5}&
  2016 &
  Food-11 &
  Generic &
  16643(11) &
  Other food datasets 
\\ \hline
  Farinella et al. \cite{dataset13} &
  2016
 &
  UNCIT-FD1200 & Generic
   &
  4754(1200) & Acquired using smartphone
    
   \\ \hline
    Jaclyn Rich et al.  \cite{dataset19}  &
  2016
 &
  Instagram 800k & Generic
   &
  808,964(43) & Social Media
   \\ \hline
   Liang et al. \cite{dataset20}&
  2017
 &
  ECUSTFD & Generic
   &
  2978(19) & Acquired using smartphone
  
   \\ \hline
Güngör et al. \cite{dataset16}&
  2017&
  Turkish-Foods-15 &
  Turkish Dishes &
  7500/15 &
 \begin{tabular}[c]{@{}l@{}}  Collected from\\ other datasets \end{tabular} 
   \\ \hline
Pandey et al, \cite{d23}&2017&
\begin{tabular}[c]{@{}l@{}}  Indian Food\\ Database\end{tabular} &
  Indian Foods &
  5000(50) &
  Downloaded from web \\ \hline
 
Termritthikun et al. \cite{d31}&2017&
  THFood-50 &
  Thai Foods &
  700/50 &
  Downloaded from web 
   \\ \hline
  Ciocca et al. \cite{dataset14} &
  2017
 &
  FOOD524DB & Generic
   &
  247,636(524) & Existing food database
 
   \\ \hline
   Hou et al. \cite{dataset7}&
  2017&
  VegFru &
\begin{tabular}[c]{@{}l@{}}  Generic (Fruit and VEG)\end{tabular} &
  160,731(292) &
  Collected from search engine
   \\ \hline
  Waltner et al. \cite{ma6}&
  2017&
  FruitVeg-81 &
 \begin{tabular}[c]{@{}l@{}} Generic (Fruit and VEG)\end{tabular} &
  15,630(81) &
   \begin{tabular}[c]{@{}l@{}}Collected using\\ mobile phone\end{tabular}  
   \\ \hline
Muresan et al. \cite{dataset6}&
  2018&
  Generic (Fruits 360 Dataset) &
  Fruit  Dataset &
  71,125(103) &
  Camera 
   \\ \hline
   Qing Yu et al. \cite{dataset15}&
   2018
 &
  FLD-469 & Japanese Foods
   &
  209,700(469) & Smart Phone camera
   
   \\ \hline
   Kaur et al. \cite{dataset8}&
  2019 &
  FoodX-251 &
  Generic &
  158,000(251) &
  Collected from web 
   \\ \hline
Ghalib et al. \cite{d42}&
  2020&
\begin{tabular}[c]{@{}l@{}}Pakistani Food\\ Dataset\end{tabular} &
  Pakistani Dishes &
  4928(100) &
  Crawled from web  
   \\ \hline

 Narayanan et al. \cite{dataset23}
&
 &
\begin{tabular}[c]{@{}l@{}}AI-Crowd \end{tabular} & Swiss Foods
   &
  25389 &Volunteer Users
   &
   &
   &
 
   \\ \hline

  Bolaños M. et al. \cite{dataset21}&
     2016
 &
  EgocentricFood & Generic
   &
  5038(9) & \begin{tabular}[c]{@{}l@{}}Taken by a wearable\\ egocentric vision camera\end{tabular}

   \\ \hline
    E. Aguilar et al. \cite{dataset22}&
   2019
 &
  MAFood-121 & Spanish Foods
   &
    21,175
   &
    Google search engine
   \\ \hline

\end{tabular}%
}
\end{table*}

\begin{figure*}[!htbp]
\centering
\includegraphics[width=7in]{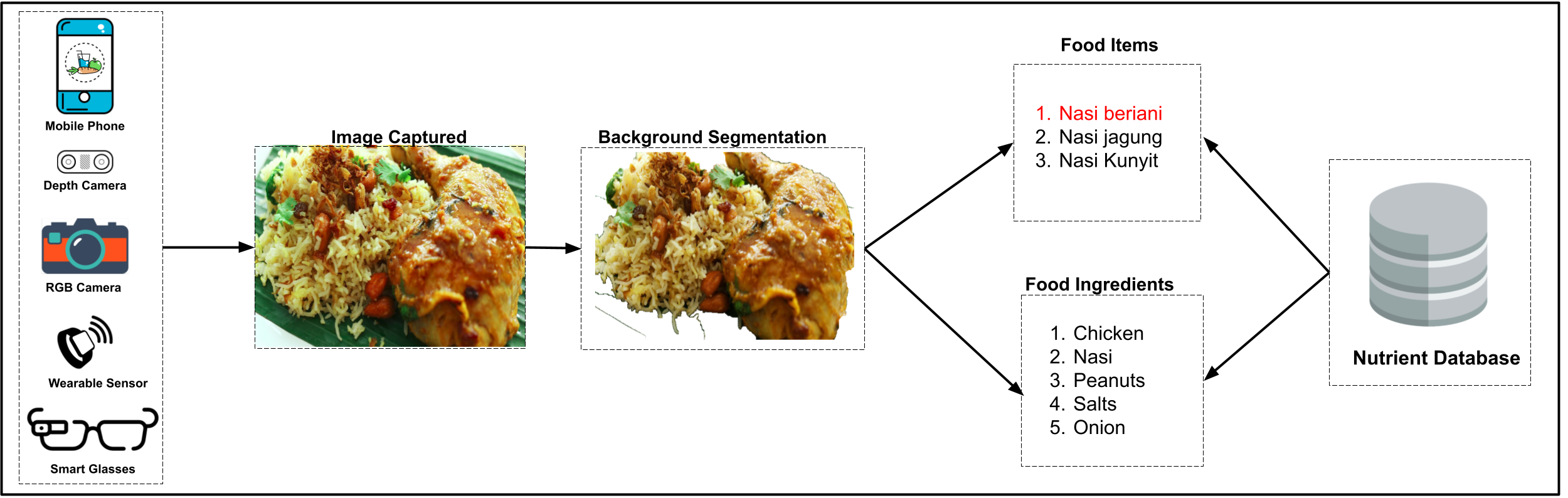}
\caption{System Flow}
\label{fig_systemflow}
\end{figure*}

\begin{figure*}[!htbp]
\centering
\includegraphics[width=7in]{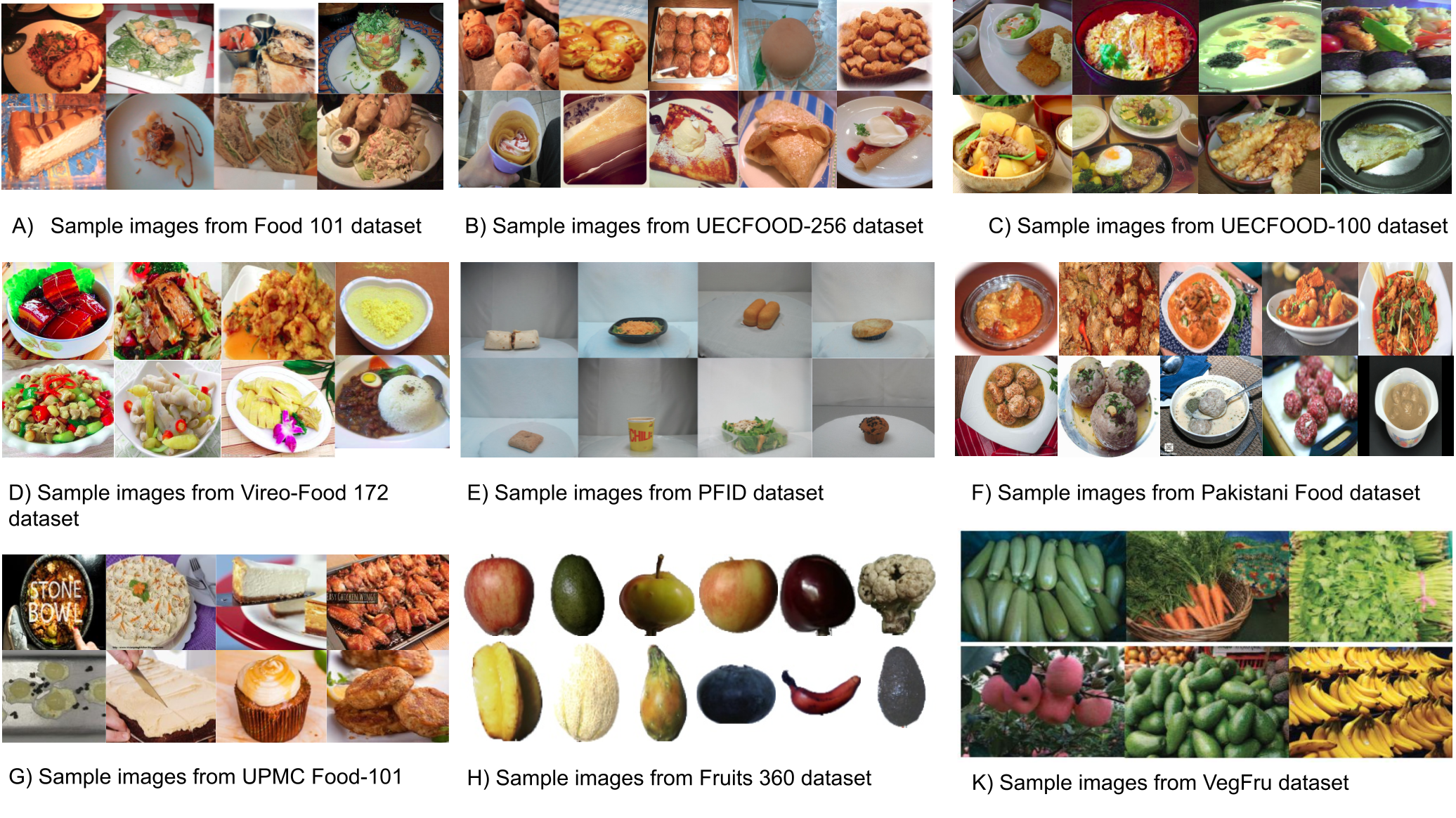}
\caption{Sample images from few food datasets}
\label{fig_sampleimages}
\end{figure*}
\par Therefore, it is evident from the survey that there is an immense need for broad and generic food datasets for better food recognition and enhanced performance. This necessity is because region-specific food items or datasets with fewer food categories can undermine the accuracy and performance of classification and extraction methods.
\section{Representation of Food Images}
Feature extraction plays a vital role in automated food recognition applications due to its noticeable impact on the recognition efficiency of an employed system. Feature extractors methods extract different food image representations. The process of feature extraction involves the identification of visual characteristics like colour, shape, and texture. The main objective of feature extraction is to reduced dimensionality space \cite{8} and extract more manageable groups from raw vectors of food images. 
\par Moreover, selecting the right set of features ensures that relevant information is extracted from input images to perform the desired task. We categorized the feature extraction techniques into two main types: Hand-crafted and Deep visual features. The term ‘handcrafted’ refers to identifying relevant feature vectors of appropriate objects such as shape, colour, and texture. In contrast to that, the deep model provides state-of-the-art performance due to automatic feature extraction through series of connected layers. For this reason, recent studies have adopted combinations of both hand-crafted and deep visual features for food image representation.   
\subsection{Handcrafted Features}
Existing literature exhibits a large number of methods to employ manually designed or handcrafted features. Handcrafted features are properties obtained through algorithms using help from information available in the image. Figure \ref{fig_handcraftedfeatures}  categorizes the handcrafted feature extraction methods. In the scenario of food image recognition, there is a variation among different food types in terms of texture, shape, and colour.
\par The term ‘texture’ refers to homogenous visual patterns that do not result from single colours such as sky and water [7]. Texture features usually consist of regularity, coarseness, and frequency. Texture-based characteristics are classified into two classes, namely statistical model and transform-based. Similarly, shape features attempt to quantify shape in ways that agree with human intuition or aid in perception based on relative proximity to well-known shapes. Based on the analysis, these shapes can either be declared perceptually similar to human perception or different. Also, extracted features should remain consistent concerning rotation, location, and scaling (changing the object size) of an image.
Unlike shape and texture features, colour features are prevalent for image retrieval and image classification because of their invariant property concerning image translation, scaling, and rotation. The key items of the colour features extraction process are colour quantization and colour space. Therefore, the resulting histogram is only discriminative, when it projects the input image is to the appropriate colour space. Different methods like; HSV, CIELab, RGB, normalized RGB, opponent colour spaces, colour k-means clustering, a bag of colour features, colour patches, and colour-based kernel, are widely employed for food classification. Although, the colour features from the food images distinguish between different food items. However, due to intra-class similarity, these features alone are not enough to accurately classify food images. For this reason, most of the researchers have used colour features in combination with other feature extraction methods.
\par Hoashi et al. \cite{hf1} employed Bag of features, colour histogram, Gabor features, and gradient histogram with multiple kernel learning for automatic food recognition of 85 different food categories. Similarly, Yang et al., \cite{hf2} deal with pairwise statistics between local features for food recognition purposes using the PFID dataset. For real-time food image recognition Kawano and Yanai et al., 2014 \cite{hf11} utilized handcrafted features like colour, HOG, and fisher vector. Moreover, the cloud-based food recognition method proposed by Pouladzadeh et al., 2015 \cite{hf12} involves features like colour, texture, size, shape, and Gabor filter. They evaluated their framework on single food portions consisting of fruit and a single item of food. Furthermore, mobile food recognition systems proposed by Kawano and Yanai, 2013 \cite{hf6}, and Oliveira et al. 2014 \cite{hf16} also used handcrafted features like colour and texture.  Table \ref{handcraftedfeatures} summarizes the details of proposed methods that employ handcrafted features for food recognition. 
\par However, identification of food involves challenges due to varying recipes and presentation styles used to prepare food all around the globe resulting in different feature-sets\cite{23}. For instance, the shape and texture of a salad containing vegetables differ from the shape, and texture of a salad containing fruits. For this reason, we should optimize the feature extraction process by extracting relevant visual information from food images. Such data is present in general information descriptors, which are a collection of visual descriptors that provide information about primary features like shape, colour, texture, and so forth. Some important descriptors used in existing studies include Gabor Filter, Local Binary Patterns (LBP), Scale-invariant Feature Transform (SIFT), and colour information to extract features of food images \cite{hf4}. These descriptors can be applied individually or in combination with other descriptors for enhanced accuracy.

\par  Nonetheless, feature selection remains a complex task for food types that involve mixed and prepared foods. Such food items are difficult to identify and are not easily separable due to the proximity of ingredients in terms of colour and texture features. In contrast, the evolution of deep learning methods has remarkably reduced the use of handcrafted features. It is due to their superior performance for both food categorization and ingredient detection tasks. However, handcrafted methods for feature extraction may still serve as the foundation for automated food recognition systems in the future.
\begin{table*}[!ht]
\centering
\caption{Handcrafted features}
\label{handcraftedfeatures}
\resizebox{\textwidth}{!}{%
\begin{tabular}{lllll}
\hline
Reference & Year&
  Visual Features &
  Dataset&
  Recognition type \\ \hline
\begin{tabular}[c]{@{}l@{}}Hoashi et   al. \cite{hf1}\end{tabular}&\begin{tabular}[c]{@{}l@{}}2010\end{tabular} &
  \begin{tabular}[c]{@{}l@{}}Bag-of-features(BOF),  \\ Color histogram, Gabor features,\\ and gradient histogram \\ with Multiple Kernel learning.\end{tabular} &
  \begin{tabular}[c]{@{}l@{}}Used for recognition \\ of 85 food categories\end{tabular} &
  \begin{tabular}[c]{@{}l@{}}Automatic \\ food recognition\end{tabular} \\ \hline
\begin{tabular}[c]{@{}l@{}}Yang et al. \cite{hf2}\end{tabular}&\begin{tabular}[c]{@{}l@{}}2010\end{tabular} &
  \begin{tabular}[c]{@{}l@{}}Deals with \\ pair wise statistics \\ between local features\end{tabular} &
  \begin{tabular}[c]{@{}l@{}}Pittsburgh \\ Food Image Dataset \\ (PFID)\end{tabular} &
  Food recognition \\ \hline
kong and Tan \cite{hf3} &\begin{tabular}[c]{@{}l@{}}2011\end{tabular} &
  \begin{tabular}[c]{@{}l@{}}SIFT,   \\ Guassian Region detector\end{tabular} &
  \begin{tabular}[c]{@{}l@{}}Pittsburgh Food Image Dataset (PFID) \\ and dataset consisting of \\ food images collected   \\ from local restaurants.\end{tabular} &
  \begin{tabular}[c]{@{}l@{}}Regular shaped \\ foods recognition\end{tabular} \\ \hline
Bosh et al. \cite{hf4} &\begin{tabular}[c]{@{}l@{}}2011\end{tabular}& 
  \begin{tabular}[c]{@{}l@{}}Global feature classes: texture and color\\ Local features: local entropy color, local color,\\ Garbor filter, SIFT, Haar, Daisy   descriptor, \\ Steerable filters and Tamura perceptual filter\end{tabular} &
  \begin{tabular}[c]{@{}l@{}}Database consisting of\\ food images collected under controlled \\ conditions, from  nutritional studies \\ conducted at \\ Prudue University \cite{hf14} \end{tabular} &
  \begin{tabular}[c]{@{}l@{}}Food recognition \\ and quantification\end{tabular} \\ \hline
  \begin{tabular}[c]{@{}l@{}}Zhang   et. \cite{hf15}\end{tabular}&\begin{tabular}[c]{@{}l@{}}2011\end{tabular} &
  Color, SIFT, Shape, RGB histograms &
  \begin{tabular}[c]{@{}l@{}}Dataset came from online sources, \\ which includes three types of   \\ cuisines, two dishes per cuisines were \\ represented by 76 images\end{tabular} &
  \begin{tabular}[c]{@{}l@{}}Classification \\ of cuisines\end{tabular} \\ \hline
\begin{tabular}[c]{@{}l@{}}Matsuda et al. \cite{hf5}\end{tabular} &\begin{tabular}[c]{@{}l@{}}2012\end{tabular}&
  \begin{tabular}[c]{@{}l@{}}Gabor texture features, \\ Histogram of Oriented Gradient (HoG), \\ Bag-of-features of SIFT and CSFIT \\ with Spatial pyramid.\end{tabular} &
  \begin{tabular}[c]{@{}l@{}}Food image dataset \\ containing 100 different \\ food categories.\end{tabular} &
  \begin{tabular}[c]{@{}l@{}}Multiple \\ food images\\ recognition\end{tabular} \\ \hline
\begin{tabular}[c]{@{}l@{}}Kawano and Yanai \cite{hf6} \end{tabular} &\begin{tabular}[c]{@{}l@{}}2013\end{tabular}&
  \begin{tabular}[c]{@{}l@{}}Bag-of-features and Color histogram, HOG \\ patch descriptor and color patch descriptor.\end{tabular} &
  - &
  \begin{tabular}[c]{@{}l@{}}Mobile food   \\ recognition\end{tabular} \\ \hline
\begin{tabular}[c]{@{}l@{}}Anthimopoulos et al. \cite{hf7} \end{tabular} &\begin{tabular}[c]{@{}l@{}}2014\end{tabular}&
  \begin{tabular}[c]{@{}l@{}}Bag-of-features,  \\ SIFT and HSV color space\end{tabular} &
  \begin{tabular}[c]{@{}l@{}}Visual dataset consisting of 5000   \\ food images organized \\ into 11 different classes\end{tabular} &
  \begin{tabular}[c]{@{}l@{}}Food recognition\\ system for \\ diabetic patients\end{tabular} \\ \hline
\begin{tabular}[c]{@{}l@{}}Tammachat and \\ Pantuwong \cite{hf8} \end{tabular} &\begin{tabular}[c]{@{}l@{}}2014\end{tabular} &
  \begin{tabular}[c]{@{}l@{}}Bag-of-features (BoF) , Texture \\ and Color\end{tabular} &
  \begin{tabular}[c]{@{}l@{}}Database consisting of 40 types of  \\ Thai food consisting of 100 images \\ of each food type.\end{tabular} &
  \begin{tabular}[c]{@{}l@{}}Food image \\ recognition\end{tabular} \\ \hline
\begin{tabular}[c]{@{}l@{}}Pouladzadeh et al. \cite{hf9} \end{tabular}&\begin{tabular}[c]{@{}l@{}}2014\end{tabular} &
  Graph cut, Color and Texture &
  \begin{tabular}[c]{@{}l@{}}Dataset consisting of 15 different \\ categories of fruits and food.\end{tabular} &
  \begin{tabular}[c]{@{}l@{}}Food image \\ recognition \\ for calorie estimation\end{tabular} \\ \hline
\begin{tabular}[c]{@{}l@{}}He et   al. \cite{hf10} \end{tabular} & &\begin{tabular}[c]{@{}l@{}}2014\end{tabular}
  \begin{tabular}[c]{@{}l@{}}Color, Texture, DCD, SCD, SIFT, \\  MDSIFT, EFD and GFD\end{tabular} &
  \begin{tabular}[c]{@{}l@{}}Food image dataset containing 1453\\  images\end{tabular} &
  \begin{tabular}[c]{@{}l@{}}Food image \\ analysis\end{tabular} \\ \hline
\begin{tabular}[c]{@{}l@{}}Kawano and Yanai \cite{hf11}\end{tabular} &\begin{tabular}[c]{@{}l@{}}2014\end{tabular}& 
  Color, HOG and Fisher Vector &
  UECFOOD-256 food image dataset &
  \begin{tabular}[c]{@{}l@{}}Real-time food \\ image recognition\end{tabular} \\ \hline
  \begin{tabular}[c]{@{}l@{}}Oliveira   et al. \cite{hf16}\end{tabular} &\begin{tabular}[c]{@{}l@{}}2014\end{tabular}&
  Color, Texture &
  \begin{tabular}[c]{@{}l@{}}Images were gathered using  \\  mobile’s camera\end{tabular} &
  \begin{tabular}[c]{@{}l@{}}Mobile Food\\  Recognition\end{tabular} \\ \hline
  \begin{tabular}[c]{@{}l@{}}Pouladzadeh et al. \cite{hf12}\end{tabular} &\begin{tabular}[c]{@{}l@{}}2015\end{tabular}&
  Color, Texture, Size, Shape, Gabor   filter &
  \begin{tabular}[c]{@{}l@{}}System was tested on single food  \\ portions consisting of fruits and \\ single piece of food. 100 images were\\ chosen for training and 100 for\\  testing purposes.\end{tabular} &
  \begin{tabular}[c]{@{}l@{}}Cloud-based\\ food recognition.\end{tabular} \\ \hline
\begin{tabular}[c]{@{}l@{}}Farinella et al. \cite{hf13}\end{tabular} &\begin{tabular}[c]{@{}l@{}}2016\end{tabular} &
  SIFT, Bag of Textons, PRICoLBP &
  UNICT-FD1200 dataset. &
  \begin{tabular}[c]{@{}l@{}}Food image \\ recognition\end{tabular} \\ \hline
\end{tabular}
}
\end{table*}
\subsection{Deep Visual Features}
Recently, deep learning techniques have gained immense attention due to their superior performance for image recognition and classification. The deep learning approach is a sub-type of machine learning, and it trains more constructive neural networks. The vital operation of deep learning approaches includes automatic feature extraction through the sequence of connected layers leading up to a fully connected layer, which is eventually responsible for classification. Moreover, in contrast to conventional methods, deep learning techniques show outstanding performance while processing large datasets and have excellent classification potential \cite{11}\cite{12}.
\par Deep learning methods like convolutional neural networks (CNN) \cite{13}, Deep Convolutional Neural Networks (DCNN) \cite{d18}, Inception-v3 \cite{15} and Ensemble net are implemented by existing food recognition methods for feature extraction. Convolutional Neural network is one of the widely used deep learning techniques in the area of computer vision due to its impressive learning ability regarding visual data and achieves higher accuracy in contrast to other conventional techniques \cite{16}.  DCNN technique gained popularity owing to its large-scale object recognition ability. It incorporates all major object recognition procedures like feature extraction, coding, and learning. Therefore, DCNN is an adaptive approach for estimating adequate feature representation for datasets \cite{17}. Similarly, Inception-v3 is also a new deep convolutional neural network technique introduced by Google. It is composed of small inception modules which are capable of producing very deep networks. As a result, this model has proved to have higher accuracy, decreased number of parameters, and computational cost in contrast to other existing models. Likewise, Ensemble Net is a deep CNN-based architecture and is a suitable method for extracting features. It is due to the outstanding performance of CNN feature descriptors as compare to handcrafted features.

\begin{table*}[!ht]
\centering
\caption{Deep visual features}
\label{deepvisualfeatures}
\resizebox{\textwidth}{!}{%
\begin{tabular}{lllll}
\hline
Reference & Year&
  Features &
  Dataset &
  Recognition type \\ \hline
  Kawano and Yanai, \cite{d26} &2014&
  Fisher Vector and DCNN &
  \begin{tabular}[c]{@{}l@{}}UECFOOD-100 and\\ 100-class food Dataset\end{tabular} &
  Food image recognition \\ \hline
Yanai and Kawano, \cite{d18} &2015&
  DCNN &
  \begin{tabular}[c]{@{}l@{}}UECFOOD-100 \\ and UECFOOD- 256\end{tabular} &
  Food image recognition \\ \hline
  Christodoulidis   et al. \cite{d24} &2015&
  CNN &
  \begin{tabular}[c]{@{}l@{}}Manually annotated dataset \\ with 573 food items\end{tabular} &
  Food recognition \\ \hline
Pouladzadeh et   al. \cite{d25} &2016&
  Graphcut and DCNN &
  \begin{tabular}[c]{@{}l@{}}Database consisting of \\ 10000 high res images\end{tabular} &
  \begin{tabular}[c]{@{}l@{}}Food recognition for \\ calorie measurement\end{tabular} \\ \hline
  Hassannejad et   al.\cite{d21} &2016&
  Inception &
  \begin{tabular}[c]{@{}l@{}}Food-101, UECFOOD-100 \\ and UECFOOD-256\end{tabular} &
  Food image recognition \\ \hline
  Liu et al. \cite{d22} &2016&
  DCNN &
  Food-101, UECFOOD-256 &
  Mobile food image recognition \\ \hline
Chen and Ngo, \cite{hclassification1} &2016&
  Arch-D &
  Chinese Foods  &
  \begin{tabular}[c]{@{}l@{}}Ingredient recognition \\ and food categorization\end{tabular} \\ \hline
  Ciocca et al. \cite{d30} &2017&
  VGG &
  UNIMIB 2016 &
  Food recognition \\ \hline
Termritthikun et   al. \cite{d31} &2017&
  NU-InNet &
  THFOOD-50 &
  Food recognition \\ \hline
  
  Pandey et al. \cite{d23} &2017&
  \begin{tabular}[c]{@{}l@{}}AlexNet, GoogLeNet \\ and ResNet\end{tabular} &
  \begin{tabular}[c]{@{}l@{}}ETH Food-101 and \\ Indian Food Image Database\end{tabular} &
  Food Recognition \\ \hline
Liu et al. \cite{d20} &2018&
  GoogleNet &
  \begin{tabular}[c]{@{}l@{}}UECFOOD-100,UECFOOD-256\\  and Food-101\end{tabular} &
  \begin{tabular}[c]{@{}l@{}}Food recognition \\ for dietary assessment\end{tabular} \\ \hline

McAllister et al. \cite{d32} &2018&
  ResNet-152, GoogLeNet &
  \begin{tabular}[c]{@{}l@{}}Food 5k, Food-11,RawFooT-DB \\ and Food-101\end{tabular} &
  Food recognition \\ \hline
Martinel et al. \cite{d33} &2018&
  WISeR &
  \begin{tabular}[c]{@{}l@{}}UECFOOD-100, UECFOOD-256 \\ and Food-101\end{tabular} &
  Food recognition \\ \hline
  E. Aguilar et   al. \cite{d35} &2018&
AlexNet &
  UNIMIB2016 &
  Automatic food tray analysis \\ \hline
  S. Horiguchi et   al. \cite{d40} &2018&
  GoogleNet &
  Built their own food dataset FoodLog &
  Food image recognition \\ \hline
  Gianluigi Ciocca et al. \cite{d45} &2018&
  ResNet50 &
  Food 475 &
  \begin{tabular}[c]{@{}l@{}}Food image recognition \\ and classification\end{tabular} \\ \hline
B. Mandal et al. \cite{d41} &2019&
  SSGAN &
  ETH Food-101 and Indian Food Dataset &
  \begin{tabular}[c]{@{}l@{}}Food Recognition \\ of Partially Labeled Data\end{tabular} \\ \hline
G.Ciocca et   al. \cite{d34} &2020&
  \begin{tabular}[c]{@{}l@{}}GoogleNet, Inception-v3,\\ MobileNet-V2 and ResNet-50\end{tabular} &
  \begin{tabular}[c]{@{}l@{}}Own dataset containing 20 different\\ food categories of fruit and   \\ vegetables.\end{tabular} &
  \begin{tabular}[c]{@{}l@{}}Food category recognition, \\ Food state recognition\end{tabular} \\ \hline

L. Jiang et al. \cite{d36} &2020&
  VGGNet &
  \begin{tabular}[c]{@{}l@{}}UECFOOD-100, UECFOOD-256 and \\ introduced new \\ dataset based on FOOD-101.\end{tabular} &
  Food recognition and dietary assesment \\ \hline
C. Liu et al. \cite{d37} &2020&
  VGGNet, ResNet &
  Vireo-Food 172 &
  Food ingredient recognition \\ \hline
H. Liang et al. \cite{d38} &2020&
   &
  ChineseFoodNet and Vireo-Food 172 &
  Chinese food recognition \\ \hline
H. Zhao et al. \cite{d39} &2020&
  VGGNet, ResNet and DenseNet &
  UECFOOD-256 and Food-101 &
  Mobile food recognition \\ \hline

G. A. Tahir and C. K. Loo \cite{d42} &2020&
  \begin{tabular}[c]{@{}l@{}}ResNet-50, DenseNet201 \\ and InceptionResNet-V2\end{tabular} &
  \begin{tabular}[c]{@{}l@{}}Pakistani Food Dataset, UECFOOD-100,\\   UECFOOD-256, FOOD-101 and PFID\end{tabular} &
  Food recognition \\ \hline
C. S. Won \cite{d43} &2020&
  ResNet50 &
  \begin{tabular}[c]{@{}l@{}}UECFOOD-256, Food-101 \\ and Vireo-Food 172\end{tabular} &
  \begin{tabular}[c]{@{}l@{}}Fine grained \\ Food image recognition\end{tabular} \\ \hline
Zhidong Shen et al. \cite{d44} &2020&
  Inception-v3, Inception-v4&
  \begin{tabular}[c]{@{}l@{}}Dataset was created including\\ hundreds and thousands of images \\ of   several food categories\end{tabular} &
  \begin{tabular}[c]{@{}l@{}}Food recognition \\ and nutrition estimation\end{tabular} \\ \hline

\end{tabular}
}
\end{table*}
\par Asymmetric multi-task CNN and spatial pyramid CNN \cite{18} provides highly discriminative image representations. Jing et al. \cite{hclassification1} proposed ARCH-D architecture for multi-class multilabel food recognition, and their model provides feature vectors for both food category and ingredient recognition. Although the feature vectors from multi-scale multi-view deep network \cite{19}   has a very high dimension however they were successful in achieving state-of-art performance. Ghalib et al. \cite{d42} proposed ARCIKELM for open-ended learning. They have employed InceptionResnetV2 for feature extraction due to their superior performance over other deep feature extraction methods such as ResNet-50 and DenseNet201. Table \ref{deepvisualfeatures} further provides a brief description of deep visual features.
\begin{figure}[!htbp]
\centering
\includegraphics[width=3.5in]{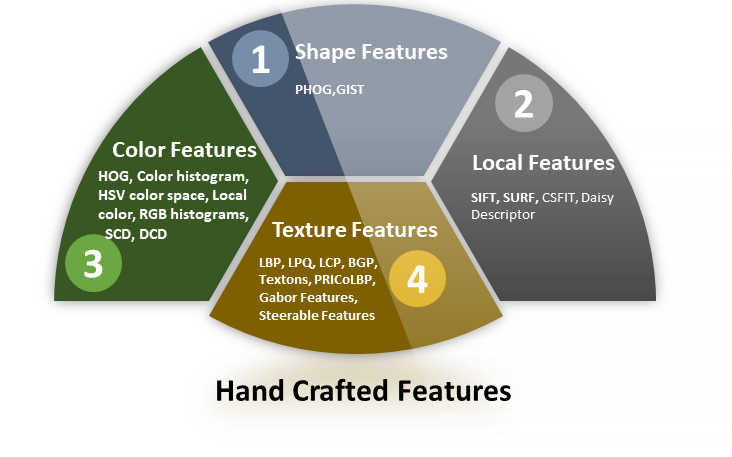}
\caption{Handcrafted feature extraction methods}
\label{fig_handcraftedfeatures}
\end{figure}

\section{Food Category Classification}
The primary requirement of any food recognition system is accurate identification and recognition of food components in the meal.Therefore, robust and precise food classification methods are crucial for several health-related applications like automated dietary assessment, calorie estimation, and food journals. Image classification refers to a machine learning technique that associates a set of unspecified objects with a subset (class) learned by the classifier during the training phase. In the scenario of food image classification, food images are used as input data to train the classifier. Hence, an ideal classifier must recognize any food category explicitly included during the learning phase. The accuracy of a classifier mainly depends on the quantity and quality of images, as there are several variations in food images such as rotation, distortion, lightning distribution, and so forth. In this section, we discussed classification techniques used by traditional approaches which use handcrafted features. Following that, we analyzed state-of-the-art deep learning models for food recognition. 
\subsection{Traditional Machine Learning Methods}
Major classifiers used by several traditional approaches in the domain of food image recognition include Support Vector Machines (SVM)\cite{hf1}, Multiple Kernel Learning (MKL) \cite{hf1} and K-Nearest Neighbor (KNN)\cite{hclassification1}.  It is due to their outstanding performance as compared to other classification methods. 
\par Food recognition method proposed by \cite{hclassification2} employs colour, SIFT, and texture features to train KNN classifier. In contrast to SVM, KNN achieved higher classification accuracy i.e. 70\%, whereas the accuracy of the SVM classifier was only 57\%.  Similarly, Anthimopoulos et al.,\cite{hf7} applied the Bag-of-features model using SIFT features. They also trained SVM linear image classifier to recognize 11 different food classes and acquired a classification accuracy of 78\%. 
\par Chen et al.\cite{hclassification3}, employed a multi-class SVM classifier for the identification of 50 different classes of Chinese food. It includes 100 food images in each category. However, obtained classification accuracy was 62.7\% only. They further implemented a multi-class Adaboost algorithm and enhanced their classification accuracy up to 68.3\%. Furthermore, Bejibom et al., \cite{hclassification4} used LBP, colour, SIFT, MR8, and HOG features to train an SVM image classifier. They evaluated their work on two different datasets and achieved a classification accuracy of 77.4\% on the dataset presented by \cite{hclassification3},  their classification accuracy was 51.2\% when applied to the menu-matched dataset. 
Table \ref{tab:traditionalmethodsclassficationAcuuracy} summarizes classifiers implemented by traditional classification methods along with their achieved classification accuracies.
\begin{table*}[!htbp]
\centering
\caption{Traditional machine learning methods for food category classification}
\label{tab:traditionalmethodsclassficationAcuuracy}
\resizebox{\textwidth}{!}{%
\begin{tabular}{lllll}
\hline
Reference &\multirow{2}{*}{Year} &
  \multirow{2}{*}{Classification Technique} &
  \multicolumn{2}{c}{Classification Accuracy} \\ \cline{4-5} 
 & 
   & &
  Top 1 &
  Top 5 \\ \hline
Hoashi et al. \cite{hf1} & 2010&
  Multiple Kernel Learning (MKL) &
  \begin{tabular}[c]{@{}l@{}}Own Food Dataset = 62.5\%\\ \end{tabular} &
  N/A \\ \hline
Yang et al. \cite{hf2} & 2010&
  Support Vector Machine   (SVM) &
  PFID = 78.0\% &
  N/A \\ \hline
kong and Tan \cite{hf3} & 2011&
  Multi-class SVM &
  PFID = 84\% &
  N/A \\ \hline
Bosh et al. \cite{hf4} & 2011&
  Support Vector Machine   (SVM) &
  \begin{tabular}[c]{@{}l@{}}Dataset collected   =86.1\%\\ using nutritional   \\ studies Conducted\\ at Prudue University\end{tabular} &
  N/A \\ \hline
  Zhang et al. \cite{hf15} & 2011&
  SVM regression with RBF kernel &
  \begin{tabular}[c]{@{}l@{}}Own Food Dataset= 82.9\%\\ \end{tabular} &
  N/A \\ \hline
Matsuda et al. \cite{hf5} &2012&
  \begin{tabular}[c]{@{}l@{}}Multiple Kernel Learning (MKL) \\ and   Support Vector Machine (SVM)\end{tabular} &
  Own food Dataset = 55.8\% &
  N/A \\ \hline
Kawano and Yanai \cite{hf6}&2013&
  Linear SVM and fast toookernel &
  N/A &
  81.6\% \\ \hline
Anthimopoulos et al.\cite{hf7} &2014&
  Linear SVM &
  \begin{tabular}[c]{@{}l@{}}Own Food Dataset = 78.0\%\\ \end{tabular} &
  N/A \\ \hline
\begin{tabular}[c]{@{}l@{}}Tammachat and  \\ Pantuwong,  \cite{hf8}\end{tabular} & 2014&
  Support Vector Machine   (SVM) &
  \begin{tabular}[c]{@{}l@{}}Own Food Dataset =70.0\%\\ \end{tabular} &
  N/A \\ \hline
Pouladzadeh et al. \cite{hf9} &2014&
  Support Vector Machine (SVM) &
  Own Food Dataset = 95\%  &
  N/A \\ \hline
He et al. \cite{hf10} &2014&
  \begin{tabular}[c]{@{}l@{}}K- nearest Neighbors   \\ and Vocabulary Trees\end{tabular} &
  \begin{tabular}[c]{@{}l@{}}Own Food Dataset= 64.5\%\\ \end{tabular} &
  N/A \\ \hline
Kawano and Yanai \cite{hf11}&2014&
  One-vs-rest &
  UECFOOD-256 = 50.1\% &
  UECFOOD-256 = 74.4\% \\ \hline
  Oliveira et al. \cite{hf16} &2014&
  Support Vector Machine (SVM) &
  \begin{tabular}[c]{@{}l@{}}Own Food Dataset\\ Top 3 classification  \\  achieved between 84 and 100 \%\end{tabular} &
  N/A \\ \hline
Pouladzadeh et al. \cite{hf12} &2015&
  Cloud-based Support Vector Machine &
  \begin{tabular}[c]{@{}l@{}}Own Food Dataset = 94.5\%\\ \end{tabular} &
  N/A \\ \hline
Farinella et al. \cite{hf13} &2016&
  Support Vector Machine (SVM) &
  UNICT-FD1200 =75.74\% &
  UNICT-FD1200 = 85.68\% \\ \hline
\end{tabular}
}
\end{table*}

\subsection{Deep Learning Models}
Deep learning approaches have gained significant attention in the field of food recognition. It is due to their exceptional classification performance in comparison to traditional approaches [56] [57]. Convolutional Neural Network (CNN), Deep Convolutional Neural Network (DCNN), Ensemble Net, and Inception-v3 are some of the most prominent techniques used by existing methods for food image recognition purposes. 
\par Yanai and Kawano \cite{d26} employed a deep convolutional neural network (DCNN) on three food datasets: Food-101, UECFOOD-256, and UECFOOD-100.  They explored the effectiveness of pre-training and fine-tuning a DCNN model using 100 images from each food category obtained from each dataset. During evaluation, classification accuracy achieved was 78.77\% for UECFOOD-100, 67.57\% for UECFOOD-256 and 70.4\% for Food-101.  Similarly, the study presented by \cite{d21} implemented Inception-v3 deep network established by Google \cite{15} on the same datasets i.e. Food-101, UEC FOOD-100, and UECFOOD-256.  Classification accuracy achieved using fine-tuned model V3 was greater than classification accuracy of the fine-tuned version of DCNN i.e. 88.28\%, 81.45\%, and 76.17\% for UECFOOD-100, UECFOOD-256, and Food-101 respectively. The food recognition method proposed by \cite{d22} implemented CNN based approach using the Inception model on the same three datasets.
\par Classification accuracy achieved was 77.4\%, 76.3\% and 54.7\% for UECFOOD-100, UECFOOD-256 and Food-101 respectively. Table \ref{classification Accuracies of deep models} provides the overview of existing food recognition methods based on deep learning approaches and their classification performance.

\begin{table*}[!htbp]

\centering
\caption{Deep learning models for food category classification}
\label{classification Accuracies of deep models}
\resizebox{\textwidth}{!}{%
\begin{tabular}{lllll}
\hline
Reference &\multicolumn{1}{c}{Year}&
  Classification Technique &
  \multicolumn{2}{c}{Classification Performance} \\ \cline{4-5} 
\multicolumn{1}{c}{} &
  \multicolumn{1}{c}{} &
  \multicolumn{1}{c}{} &
  \multicolumn{1}{l}{Top1} &
  \multicolumn{1}{l}{Top 5} \\ \hline

Yanai and Kawano, \cite{d18} &2015&
  DCNN &
  \begin{tabular}[c]{@{}l@{}}UECFOOD-100 =78.8\%\\ UECFOOD-256 =67.6\%\end{tabular} &
  N/A \\ \hline
  Christodoulidis et al. \cite{d24} &2015&
  DCNN &
  Own   dataset = 84.9\% &
  N/A \\ \hline
  Chen and Ngo \cite{hclassification1} &2016&
  DCNN &
   &
   \\ \hline
Pouladzadeh et al. \cite{d25} &2016&
  DCNN + Graph cut &
  Own dataset =99\% &
  N/A \\ \hline
  Hassannejad et al. \cite{d21} &2016&
  DCNN &
  \begin{tabular}[c]{@{}l@{}}ETH Food-101 = 88.3\%\\ UECFOOD-100 = 81.5\%\\ UECFOOD-256 = 76.2\%\end{tabular} &
  \begin{tabular}[c]{@{}l@{}}ETH Food-101 = 96.9\%\\ UECFOOD-100 = 97.3\%\\ UECFOOD-256 = 92.6\%\end{tabular} \\ \hline
Liu et al. \cite{d22} &2016&
  CNN &
  \begin{tabular}[c]{@{}l@{}}UECFOOD-100   = 76.3\%\\ Food-101   =77.4\%\end{tabular} &
  \begin{tabular}[c]{@{}l@{}}UECFOOD-100   =94.6\%\\ Food-101   = 93.7\%\end{tabular} \\ \hline
  Pandey et al. \cite{d23} &2017&
  Ensemble Net  &
  \begin{tabular}[c]{@{}l@{}}ETH-Food101 = 72.1\%\\ Indian Food =73.5\%\\ Database\end{tabular} &
  \begin{tabular}[c]{@{}l@{}}ETH-Food101 = 91.6\%\\ Indian Food =94.4\%\\ Database\end{tabular} \\ \hline
  Ciocca et al. \cite{d30} &2017&
  CNN &
  UNIMIB   2016 = 78.3\% &
  N/A \\ \hline
Termritthikun et al. \cite{d31} &2017&
  CNN &
  THFOOD-50 = 69.8\% &
  THFOOD-50 = 92.3\% \\ \hline
McAllister et al. \cite{d32} &2018&
  \begin{tabular}[c]{@{}l@{}}CNN+ANN+SVM+  \\  Random Forest\end{tabular} &
  \begin{tabular}[c]{@{}l@{}}Food-5K   =99.4\%\\ Food-11   = 91.3\%\\ RawFooT-DB   = 99.3\%\\ Food-101   = 65.0\%\end{tabular} &
  N/A \\ \hline
Liu et al. \cite{d20} &2018&
  DCNN &
  \begin{tabular}[c]{@{}l@{}}UECFOOD-256 = 54.5\%\\ UECFOOD-100 = 77.5\%\\ Food   101 = 77.0\%\end{tabular} &
  \begin{tabular}[c]{@{}l@{}}UECFOOD-256 = 81.8\%\\ UECFOOD-100 = 95.2\%\\ Food   101 = 94.0\%\end{tabular} \\ \hline

Martinel et al. \cite{d33} &2018&
  DNN &
  \begin{tabular}[c]{@{}l@{}}UECFOOD-100 =89.6\%\\ UECFOOD-256 =83.2\%\\ Food-101 = 90.3\%\end{tabular} &
  \begin{tabular}[c]{@{}l@{}}UECFOOD-100 =99.2\%\\ UECFOOD-256 =95.5\%\\ Food-101 = 98.7\%\end{tabular} \\ \hline
E. Aguilar et al. \cite{d35} &2018&
  CNN+SVM &
  UNIMIB   2016 = 90.0\% &
  N/A \\ \hline
  Gianluigi Ciocca et al. \cite{d45} &2018&
  CNN &
  Food-475 = 81.6\% &
  Food-475 = 95.5\% \\ \hline
  S. Horiguchi et   al. \cite{d40} &2018&
  \begin{tabular}[c]{@{}l@{}}Sequential Personalized Classifier \\ (SPC) with   \\ fixed-class and incremental \\ classification\end{tabular} &
  \begin{tabular}[c]{@{}l@{}}FoodLog = 40.2\%\\ (t251-t300)\end{tabular} &
  \begin{tabular}[c]{@{}l@{}}FoodLog = 56.6\%\\ (t251-t300)\end{tabular} \\ \hline
  B. Mandal et al. \cite{d41} &2019&
  Generative Adversarial Network  &
  \begin{tabular}[c]{@{}l@{}}ETH   Food-101 =  75.3\%\\ IndianFood Database = 85.3\%\end{tabular} &
  \begin{tabular}[c]{@{}l@{}}ETH   Food-101 =  93.3\%\\ Indian   Food Database = 95.6\%\end{tabular} \\ \hline
 Aguilar-Torres et al. \cite{lr2} &2019&
  CNN based on ResNet-50 &
 \begin{tabular}[c]{@{}l@{}} MAFood-121 = 81.62\%\end{tabular} &
  N/A \\ \hline
   Kaiz Merchant and Yash Pande \cite{lr5} &2019&
  Inception V3 &
  ETHZ Food-101 =70.0\% &
  N/A \\ \hline
Mezgec, S. et al. \cite{lr8} &2019&
  Deep Learning &
 Own Food dataset = 93\%&
  N/A \\ \hline

L. Jiang et al. \cite{d36} &2020&
  DCNN (Faster R-CNN) &
  FOOD20-with-bbx = 71.7\% &
  FOOD20-with-bbx = 93.1\% \\ \hline
C. Liu et al.,2020 \cite{d37} &
   &
   &
   \\ \hline
H. Zhao et al. \cite{d39} &2020&
  JDNet &
  \begin{tabular}[c]{@{}l@{}}UECFOOD-256   = 84.0\%\\ FOOD-101   = 91.2\%\end{tabular} &
  \begin{tabular}[c]{@{}l@{}}UECFOOD-256   = 96.2\%\\ FOOD-101   = 98.8\%\end{tabular} \\ \hline

G. A. Tahir and C. K. Loo \cite{d42} &2020&
  \begin{tabular}[c]{@{}l@{}}Adaptive Reduced Class \\ Incremental Kernel Extreme  \\ Learning Machine (ARCIKELM)\end{tabular} &
  \begin{tabular}[c]{@{}l@{}}Food-101 = 87.3\%\\ UECFOOD-100 = 88.7\%\\ UECFOOD-256= 76.51\%\\ PFID = 100\%\\ Pakistani Food = 74.8\%\end{tabular} &
  N/A \\ \hline
C. S. Won \cite{d43} &2020&
  Three-scale   CNN &
  \begin{tabular}[c]{@{}l@{}}UECFOOD-256   = 74.1\%\\ Food   101 = 88.8\%\\ Vireo-Food 172   = 91.3\%\end{tabular} &
  \begin{tabular}[c]{@{}l@{}}UECFOOD-256 = 93.2\%\\ Food-101 = 98.1\%   \\ Vireo-Food 172 = 98.9\%\end{tabular} \\ \hline
Zhidong Shen et al. \cite{d44} &2020&
  CNN &
  Own dataset =85.0\% &
  N/A \\ \hline

 Jiangpeng He et al. \cite{lr3} &2020&
  18 layer ResNet &
  Own dataset =88.67\% &
  N/A \\ \hline
  Eduardo Aguilar et al. \cite{lr4} &2020&
  CNN &
  Own dataset =88.67\% &
  N/A \\ \hline
 Dario Ortega Anderez et al. \cite{lr10} &2020&
 CNN &
  Own dataset =97.10\% &
  N/A \\ \hline
    G. Song et al. \cite{lr15} &2020&
 CNN &
  Web crawled dataset =56.47\% &
  Web crawled dataset =60.33\\ \hline
  
    Limei Xiao et al. \cite{lr7} &2021&
 CNN &
  Own dataset =97.42\% &
  N/A \\ \hline
  
 Lixi Deng et al. \cite{lr12} &2021&
ResNet-50 &
School lunch dataset =95.3\% &
 N/A \\ \hline
\end{tabular}
}
\end{table*}

\section{Food Ingredient Classification}
Over the past few years, nutritional awareness among people is increasing due to their intolerance towards certain types of food, mild or severe obesity problems, or simply because of interest in maintaining a healthy diet. This rise in nutritional awareness has also stirred a shift in the technological domain, as several mobile applications facilitate people in keeping track of their diet. However, such applications hardly offer features for automated food ingredient recognition. 
\par For this purpose, several proposed models use multi-label learning for food ingredient recognition. It \cite{ml2} can be defined as the prediction of more than one output category for each input sample. Therefore, food ingredient recognition is known as a multi-label learning problem. Marc Bolanos et al. have deployed CNN as a multi-label predictor to discover recipes in terms of the list of ingredients from food images \cite{ml3}. Similarly, Yunan Wang et al. \cite{ml4}, used multi-label learning for mixed dish recognition, as they have no distinctive boundaries among them. Therefore labelling bounding boxes for each dish is a challenging task. Another system proposed by Amaia Salvador et al., \cite{ml5} regenerates recipes from provided food images along with cooking instructions. On other hand, Jingjing Chen and Chong-Wah Ngo \cite{hclassification1} proposed deep architectures for food ingredient recognition and food categorization and evaluated their proposed system on a large Chinese food dataset with highly complex food images. 
 Moreover, food ingredient recognition is often overlooked and is a challenging task, as it requires training samples under different cooking and cutting methods for robust recognition. Therefore, methods proposed by Chen et al.\cite{lr1} and J. Chen et al. \cite{lr14} focuses around food ingredient recognition. \cite{lr1} deploys multi-relational graph convolutional network which is later evaluated on Chinese and Japanese food dataset resulting in 36.7\% for UECFOOD-100 and 48.8\% for VireoFood-172. However, \cite{lr14} proposed DCNN based method for food ingredient recognition and achieved Top1 accuracy up to 86.91\% and Top 5 accuracy up to 97.59\% for Vireo Food-251.
\par Furthermore, Table \ref{multilabel} provides brief information about accuracy scores of proposed systems along with methods and dataset used.
\begin{table*}[!htbp]
\centering
\caption{Proposed methods for food ingredient classification}
\Large
\label{multilabel}
\resizebox{\textwidth}{!}{
\begin{tabular}{lllllll}
\hline
Reference &Year&
  Dataset &
  Method &
  Recall &
  Precision &
  F1 \\ \hline
Chen et al. \cite{hclassification1}&2016 &
  Vireo-Food 172 &
  \begin{tabular}[c]{@{}l@{}}Arch-D\\ (Multi-task)\end{tabular} &
  - &
  - &
  \begin{tabular}[c]{@{}l@{}}67.17\% (Micro-F1)\\ 47.18\% (Macro-F1)\end{tabular} \\ \cline{3-7} 
 &&
  UECFOOD-100 &
  \begin{tabular}[c]{@{}l@{}}Arch-D\\ (Multi-task)\end{tabular} &
  - &
  - &
  \begin{tabular}[c]{@{}l@{}}82.06\% (Micro-F1)\\ 95.88\% (Macro-F1)\end{tabular} \\ \hline
  
Bolaños et al. \cite{ml3} &2017&
  Food-101 &
  \begin{tabular}[c]{@{}l@{}}ResNet50+\\ Ingredients 101\end{tabular} &
  73.45\% &
  88.11\% &
  80.11\% \\ \cline{3-7} 
 &&
  Recipe 5k &
  \begin{tabular}[c]{@{}l@{}}ResNet50+\\ Recipe 5k\end{tabular} &
  19.57\% &
  38.93\% &
  26.05\% \\ \cline{3-7} 
 &&
  Recipe 5k &
  \begin{tabular}[c]{@{}l@{}}Inception-v3+\\ Recipe 5k (Simplified)\end{tabular} &
  42.77\% &
  53.43\% &
  47.51\% \\ \hline  
Wang, Yunan, et al. \cite{ml4} &2019&
  Economic Rice &
  \begin{tabular}[c]{@{}l@{}}Inception-V4* + NS\\ (multi-scale)\end{tabular} &
  71.90\% &
  72.10\% &
  71.40\% \\ \cline{3-7} 
 &&
  Economic Behoon &
  \begin{tabular}[c]{@{}l@{}}Inception-V4* + NS\\ (multi-scale)\end{tabular} &
  77.60\% &
  68.50\% &
  69.70\% \\ \hline 
  
Salvador, Amaia, et al. \cite{ml5} &2019&
  Recipe 1M &
  \begin{tabular}[c]{@{}l@{}}CNN \\ Auto-Encoder\end{tabular} &
  75.47\% &
  77.13\%&
  48.61\% \\ \hline
J. Chen et al. \cite{lr14} &2021 &
  VireoFood-172 &
  \begin{tabular}[c]{@{}l@{}} DCNN\\ \end{tabular} &
  - &
  - &
  \begin{tabular}[c]{@{}l@{}}75.77\% (Micro-F1)\\ \end{tabular} \\ \cline{3-7} 
  \hline
\end{tabular}
}
\end{table*}

\section{Food Volume Estimation}
Automated food volume assessment is a convoluted task involving various challenges. Highly diverse and varying composition of food, increasing varieties of edibles, different methods of preparation are only some of the factors that need to be taken into consideration. Furthermore, the quality of pictures taken for food volume estimation also impacts the accuracy. Clear pictures taken in good lighting conditions would yield different results compared to low resolution or low-light images. Thus far, several methods have been proposed for accurate estimation of food volume ranging from simple techniques such as pixel counting to complex methods like 3D image reconstruction. They have been broadly categorized as either ‘single image view’ or ‘multi-image/video view’ methods in the subsequent sections.
\begin{figure}[!htbp]
\centering
\includegraphics[width=3.5in]{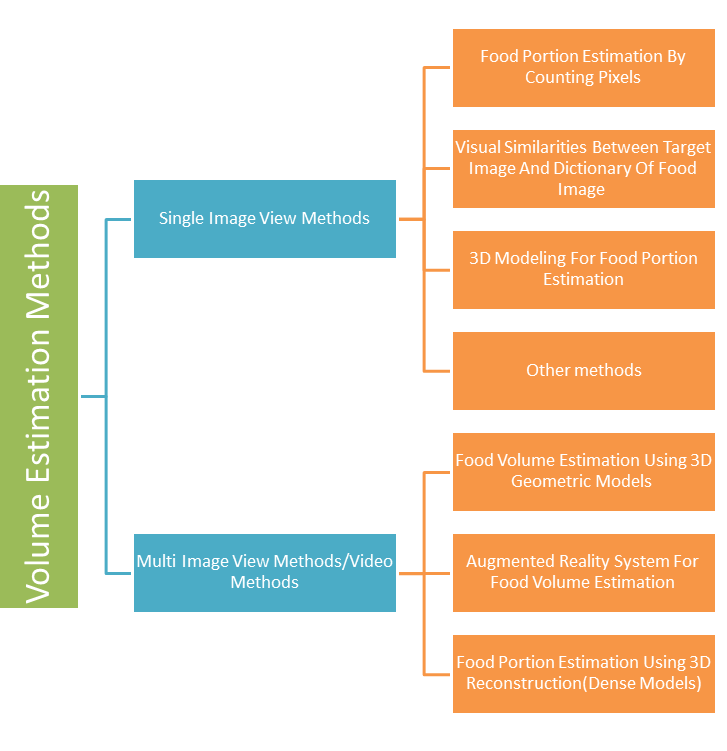}
\caption{Food Volume Estimation Methods}
\label{fig_handcraftedfeatures}
\end{figure}

\subsection{Single Image View Methods}
Single Image View Methods for food volume estimation require only a single image for food volume estimation. These methods are relatively more user-friendly as compared to ‘multi image view methods’ because they do not require multiple images from different viewpoints. However, as a trade-off, most of the ‘single view’ methods are less accurate in contrast to ‘multi view’ methods. Table \ref{singleview} summarizes single view methods for volume estimation. Following are few common methods that use ‘single view’ for food portion estimation:

\subsubsection{Food Portion Estimation by Counting Pixels}This method utilizes pixel count in each relevant image section to estimate food portion size. Studies \unskip~\cite{d44} show that these methods are less complex as compared to methods that rely on 3D modeling. Despite its simplicity, it gives good estimation of portion size, thus making caloric content and nutrient facts calculation easier.

\subsubsection{Visual Similarities Between Target Image and Dictionary of Food Image}This method estimates visual similarities between a given image and an existing food image dictionary. It is used by many existing systems today\unskip~\cite{pm2}, where the calorie and nutrient contents in the food image dictionary are defined by diet professionals to get better approximation. The method selects first `n' images from the dictionary and calculates calorie content of target image based on average calorie content of dictionary images.

\subsubsection{3d Modeling for Food Portion Estimation}This method projects a 3D model of food portions on 2D space or uses 3D geometric models for volume estimation. Generally, this method gives finer approximation in contrast to the other methods for single image view methods.

\subsubsection{Other Methods}Other methods for food portion estimation include estimating portion sizes using a ruler and adjustable wedge\unskip~\cite{datasetfv4}, Mobile Augmented Reality, Virtual reality\unskip~\cite{pm6}, visual assessment\unskip~\cite{v6} feature extraction and its matching\unskip~\cite{pm2}\unskip~\cite{hclassification4}.

\subsection{Multi-Image View or Video Methods}Multi-Image view or video methods require multiple images for food portion estimation. They are relatively more accurate as compared to single view image methods. However, multi-image methods are less user-friendly as they require multiple images from different viewpoints in order to provide better results. Table \ref{multiview} summarizes single view methods for volume estimation. Following are few methods which use multi-image view techniques for food volume estimation.

\subsubsection{Food Volume Estimation Using 3d Geometric Models}This multi-image view method uses a shape template method or 3D modeling for portion size estimation. As single shape template is not suitable for all food types, the use of geometric models with correct food classification labels and segmentation masks in the image is important to index food label to its respective class of predefined geometric models. These can be used later for finding correct parameters of selected geometric model\unskip~\cite{datasetfv4}\unskip~\cite{datasetfv3}\unskip~\cite{pm1}\unskip~\cite{pm13}\unskip~\cite{v4}.

Moreover, in 3D modeling and pose estimation, models for food are constructed in advance by using between 15 and 20 food images captured from several angles or a video sequence. Finally, food volume is estimated by registering pose from 3D model to 2D image\unskip~\cite{pm9}.

\subsubsection{Augmented Reality System for Food Volume Estimation}The use of augmented reality is also being widely used by researchers to estimate food portion size. Many systems like `Eat AR' make use of it for portion size estimation\unskip~\cite{datasetfv5}by developing prototypes to facilitate users. These prototypes generally require fiducial marker, or credit card sized objects for overlaying 3D forms. Finally, the volume of the overlaid forms is computed by using signed volume estimation algorithm for closed 3D objects.

Similarly, the `Serv Ar' augmented reality tool is used to provide guidance about food serving size\unskip~\cite{v8}. Many of these technologies are being used with object recognition methods to identify food items and determine their caloric content. Similarly, methods that use augmented reality in combination with other portion estimation techniques have enhanced accuracy and much more interactive interfaces resulting in high retention rate.

\subsubsection{Food Portion Estimation Using 3d Reconstruction (Dense Models)}Portion estimation by constructing dense 3D models usually requires multiple images or a video segment\cite{v11}. Joachim Dehais et al. \unskip~\cite{v1} have shown the use of two views for volume estimation using 3D construction. In its first stage, the system learns about the configuration of different views; followed by construction of dense 3D model to extract the volume of each individual \mbox{}\protect\newline food item placed before it. Similarly, Wen Wu et al.\unskip~\cite{pm5} study the use of fast food videos for caloric estimation. Most of these methods require images from different viewpoints, and for this reason more advance methods like 3D construction from accidental motion can be explored for food volume estimation in future.
\begin{table*}[!ht]
\centering
\scriptsize
\caption{Comparison of single-view methods for food volume estimation}
\label{singleview}
\resizebox{\textwidth}{!}{%
\begin{tabular}{p{2.5cm}lp{4cm}p{3cm}p{4cm}}
\hline
Reference &Year &
  \begin{tabular}[c]{@{}l@{}}Dataset\end{tabular} &
  \begin{tabular}[c]{@{}l@{}}Results (E: error\%)\end{tabular} &
  \begin{tabular}[c]{@{}l@{}}Technique\end{tabular} \\ \hline
S. Fang \cite{datasetfv3} &
  2015 &
  19 food items &
  E: {\textless}6\% &
  3D parameters and reference objects to compute density for estimating the weight of food item \\
Y. He\unskip~\cite{pm9} &
  2013 &
  1453 food images &
  E: 11\%(beverages) \mbox{}\protect\newline 63\% &
  "Integrated image segmentation \mbox{}\protect\newline and identification system"\\

T. Miyazaki \cite{pm2}  &
  2011 &
  6512 images &
  E:40\% &
  Linear estimation\\

Beijbom, O \cite{hclassification4}  &
  2015 &
  646 images, with 1386 tagged food items across 41 categories &
  E: 232\ensuremath{\pm}7.2 &
  Restaurant-specific food recognition considers meal as a whole entry with all of its nutrients details in DB to solve the volume estimation problem for the restaurant scenario.\\
Koichi Okamoto \cite{pm4}   &
  2016 &
  20 kinds of Japanese Foods (60 test image) &
  E:21.30\% &
 Single-image-based food calorie estimation system which uses reference objects to determine food region and quadratic curve estimation from the 2D size of foods to their calories\\

Pettitt, C \cite{v22}  &
  2016 &
  Test data from N:6 participants who completed food diary during pilot sudy by wear micro camera  &
  E:34\%  &
  Wearable micro camera in conjunction with food dairies\\

Akpa Akpro Hippocrate \cite{pm7}   &
  2016 &
  119 food images &
  E: 6.87\% &
  Image processing with cutlery\\

Jia, W. Y \unskip~\cite{pm8} &
  2012 &
  224 pictures &
  E: {\textless}10\% &
  3D location of a circular feature from a 2D image\\

Yang, Y. Q   \cite{pm6}  &
  2011 &
  72 images &
  E:-3.55\% &
  Single digital image, plate reference\\

Huang, J \unskip~\cite{pm12} &
  2015 &
  fruits(n:6) &
   &
  imaging processing\\

Yue, Y  \cite{pm14} &
  2012 &
  6 food replicas &
  E: Length (-1.18)  &
  A mathematical model based system involves a camera, circular object in a 3D space to compute food volume.\\
Zhang, W~ \cite{pm11} &
  2015 &
  15 different kinds of foods &
  85\% &
  Portion estimation by counting pixels\\
Rob Comber \cite{v6}   &
  2016 &
  6 different meals &
  " Beef (\ensuremath{E }: -13.89g, \ensuremath{\sigma  }: 5.10g), scrambled egg (\ensuremath{E }: -9.11g, \ensuremath{\sigma  }: 8.29g), Jam \mbox{}\protect\newline sponge (\ensuremath{E }: -12.31g, \ensuremath{\sigma  }: 7.03g) and fish pie (\ensuremath{E }: -12.59g, \ensuremath{\sigma  }: \mbox{}\protect\newline 5.74g). Mean: -9.58" &
  Visual Assessment\\
S. Fang  \cite{pm3} &
  2016 &
  10 objects &
   &
  "3D geometric models \mbox{}\protect\newline and depth images."\\

Godwin, S. \cite{datasetfv4} &
  2006 &
  Five portions of 9-inch cake, Seven portions of pizza, Pies were 9 or 10 inches &
  E:25\% &
  Estimated portion sizes using a ruler and the adjustable wedge\\
Hern\'{a}ndez, Teresita \cite{pm10} &
  2006 &
  101 subjects, 5 foods &
  E: 4.8\%\ensuremath{\pm}1.8\% &
  Digital photographs printed onto a poster.\\
  Yang et al. \cite{v12} &
  2021 &
  Virtual Food Dataset and Real Food Dataset (RFD) (1500 images) &
  E:$ <$ 9\% on VFD, E: 11.6\% and 20.1\% on RFD. &
Estimates volume by computing inner product between the probability vector from modified MobileNetV2 and the reference volume vector.\\

  Graikos et al. \cite{v11} &
  2021 &
  EPIC-KITCHENS and their own food video datasets &
 46.32\% average MAPE on 16 test foods and 36.90\% average MAPE on 6 combined meals. & 
Generate 3-dimensional point cloud by using depth map, segmentation mask and camera parameters. It then approximates the volume with points cloud-to-volume algorithm.\\
 Lo, F.P.W et al. \cite{v21} &
  2019 &
  Test dataset: 11 food items &
 E:15.32\%. & 
3D point cloud completion from RGB and depth images.\\
\hline 
\end{tabular}
}
\end{table*}

\begin{table*}[!ht]
\centering
\scriptsize
\caption{Comparison of multi-view methods for food volume estimation}
\label{multiview}
\resizebox{\textwidth}{!}{%
\begin{tabular}{p{2.5cm}lp{4cm}p{3cm}p{4cm}}
\hline
Reference &Year &
  \begin{tabular}[c]{@{}l@{}}Dataset\end{tabular} &
  \begin{tabular}[c]{@{}l@{}}Results (E: error\%)\end{tabular} &
  \begin{tabular}[c]{@{}l@{}}Technique\end{tabular} \\ \hline

F. Zhu \cite{v5}   &
  2010 &
  3000 images &
  E: 1\% \mbox{}\protect\newline 19 food items(97.2\%) &
  "camera calibration step and a \mbox{}\protect\newline 3D volume reconstruction step"\\

Kong, Fanyu \cite{v19}   &
  2015 &
  6 food items &
  84\%-91\% &
  Multi-View RGB images for 3D reconstruction to estimate the volume\\
Trevno, Roberto \cite{v18}  &
  2015 &
  120 students (n=120 meals; 57 breakfast + 63 lunch) &
  74\%(reliability) &
  Digital Food Imaging Analysis (DFIA)\\
Jia, W. Y \cite{v17}   &
  2014 &
  100 food samples &
  E: -2.80\% \mbox{}\protect\newline 30\% &
  ebutton for taking pictures and then portion size is calculated semi-automatically by using computer software\\

Xu, C \unskip~\cite{pm9} &
  2013 &
   &
  E:10\% &
  3D MODELLING AND POSE ESTIMATION\\

Rhyner, D \cite{v16}   &
  2016 &
  6 meals &
  85.10\% &
  Multi-View RGB images, reference card and  3D model for volume estimation\\

T. Stutz \cite{datasetfv5} &
  2014 &
  Rice, blinded servings &
  E:{\textless}33\% &
  Mobile Augmented Reality System\\

  Makhsous et al. \cite{v13} &
  2020 &
  8 food items tested &
  40\% improvement in the accuracy of volume estimation as compared to manual calculation. & 
Employs a mobile Structured Light System (SLS) to measure the food volume and portion size of a dietary intake.\\
 
 Yuan et al. \cite{v20} &
  2021 &
  Test dataset: 6 food items &
 E:0.83~5.23\%. & 
3D reconstruction from multi-view RGB images.\\
 Lo, F.P.W et al. \cite{v21} &
  2019 &
  Test dataset: 11 food items &
 E:15.32\%. & 
3D point cloud completion from RGB and depth images.\\

\hline 
\end{tabular}
}
\end{table*}

\begin{table}[!htbp]
\caption{{Summary of studies employing single image vs multiple images for volume estimation } }
\label{tw-741e08b98e5e}
\def\arraystretch{1}
\ignorespaces 
\centering 
\begin{tabular}{ll}
\hline Method & Studies\\
\hline 
Single Image &
  19\\
Multiple Images &
  10\\
\hline 
\end{tabular}\par 
\end{table}

\subsection{Strengths and Weakness of the Food Volume Estimation Methods}
Automatic food volume estimation method helps people to monitor their dietary intake suffering from chronic diseases without any expert intervention. It gives a quick result as compared to the traditional method which generally involves sending food images to the dietitian. Traditional method involves continuous involvement of dietitians which makes it unworkable for dietitians to immediately respond to a large number of patients. Conversely, automatic food volume estimation is not standardized. As there are no existing guidelines by experts which refers to the error rate of these applications. Furthermore, different volume estimation methods vary in terms of accuracy and usability. Most of these methods are classified into two categories single image view method and multiple image view method. Single view image methods are more user-friendly but accuracy is compromised as compared to multiple image view methods as it requires images from different. Therefore, standard guidelines are required for food volume estimation which should include criteria for a balanced trade between features like usability and accuracy and developed applications must be verified according to the standard guidelines.
\begin{figure}[!ht]
\centering
\includegraphics[width=3.5in]{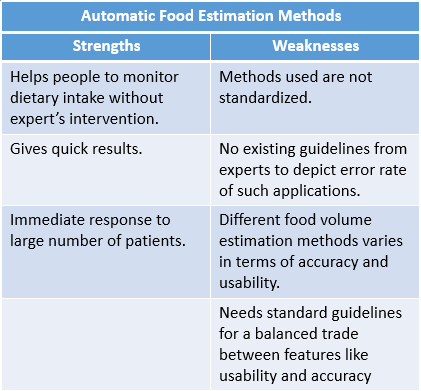}
\caption{Strengths and Weaknesses of automatic food estimation methods}
\label{fig_handcraftedfeatures}
\end{figure}

\section{Existing and Potential Applications of Vision-Based Methods for Food Recognition in Healthcare}
We summarized the core applications of vision-based methods for food recognition in context with public policy and health care.

\subsection{mhealth Apps for Dietary Assessment}
\begin{figure}[!ht]
\centering
\includegraphics[width=3.5in]{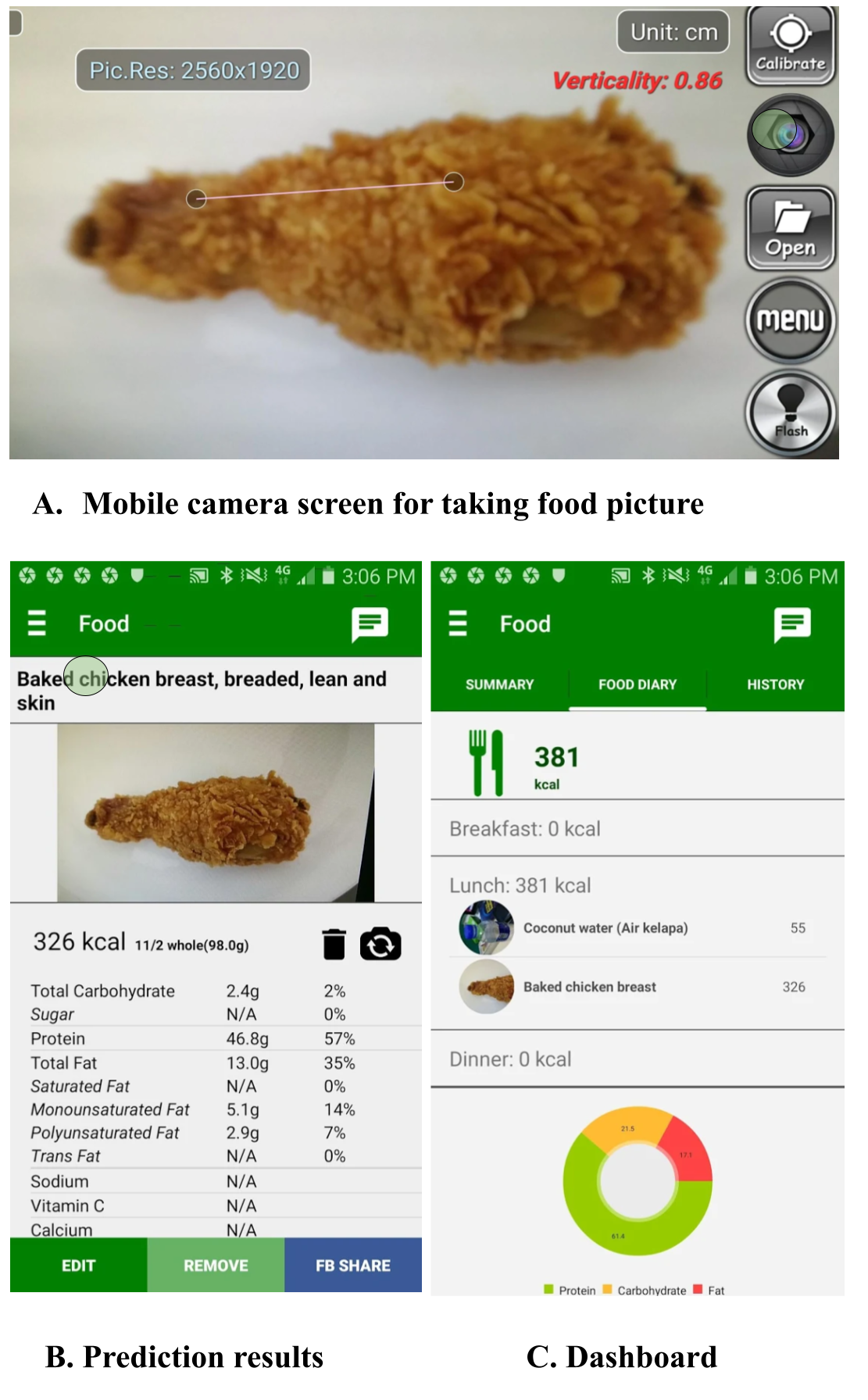}
\caption{The application provides the top 1 prediction results. This picture is taken from the study of Ghalib et al. 2020 (permission has been taken from original author)}
\label{fig_mobileapp}
\end{figure}

Today, several mobile applications have been developed to monitor diet and facilitate users to choose healthier alternatives regarding food consumption. Initially, these mobile applications were dependent on manually inputting food items by selecting from limited food databases. Therefore, such applications were not very reliable as they were prone to inaccuracies in dietary assessment, mainly extending from limited exposure to numerous food categories. With the advancement in the area of food image recognition, a large number of mHealth applications for dietary assessment use images to recognize food categories. For this purpose, existing mobile applications use different combinations of traditional and deep visual feature extraction, and classification methods for food recognition described earlier in sections III and IV. Aizawa et al., \cite{ma1} developed a mobile app ‘food log’ which uses traditional feature extraction methods like colour, Bag of Features, and SIFT; and uses an Adaboost classifier for classification purposes. Similarly, Ravi et al., \cite{ma5} proposed the ‘FoodCam’ application, which uses traditional methods for feature extraction (LBP and RGB colour features) and SVM for classification. 
Alternatively, Meyers et al., \cite{ma4} employed a deep visual technique (GoogleNet CNN model) for feature extraction and classification purposes. Similarly, the Food Tracker app proposed by Jiang et al., \cite{ma11} uses a deep convolutional neural network for feature extraction and classification.  Also, G. A. Tahir and C. K. Loo \cite{d42} utilized deep visual methods like ResNet-50, DenseNet201, and InceptionResNet-V2 for feature extraction and Adaptive Reduced Class Incremental Kernel Extreme Learning Machine (ARCIKELM) as classification method for their mobile application “My Diet Cam”. Table \ref{mobile app} summarizes existing mobile applications in terms of feature extraction and classification methods used. 
Based on these deep visual method combinations, food recognition accuracies differ for various existing mobile applications. Therefore, apps with higher food recognition and classification accuracies gain more popularity. These apps tend to ease the dietary assessment process. Figure \ref{fig_mobileapp} shows the mobile application by Ravi et al. \cite{ma5}.


\begin{table*}[!ht]
\centering
\scriptsize
\caption{Summary of feature extraction and classification methods used by existing mobile applications}
\label{mobile app}
\resizebox{\textwidth}{!}{%
\begin{tabular}{llllll}
\hline
Reference &Year &
  \begin{tabular}[c]{@{}l@{}}Application \\ Name\end{tabular} &
  \begin{tabular}[c]{@{}l@{}}Food \\ Segmentation\end{tabular} &
  \begin{tabular}[c]{@{}l@{}}Feature Extraction   \\ Method\end{tabular} &
 \begin{tabular}[c]{@{}l@{}}Classification  \\  Method\end{tabular} \\ \hline
\begin{tabular}[c]{@{}l@{}}Aizawa et al.  {\cite{ma1}}\end{tabular} & 2013  &
  FoodLog &
  No &
  \begin{tabular}[c]{@{}l@{}}Color, SIFT and \\ Bag of Features\end{tabular} &
  \begin{tabular}[c]{@{}l@{}}Adaboost \\ Classifier\end{tabular} \\ \hline
\begin{tabular}[c]{@{}l@{}}Oliveira et al. {\cite{hf16}}\end{tabular}&2014 &
  - &
  Yes &
  \begin{tabular}[c]{@{}l@{}}Color \\ and Texture\end{tabular} &
  \begin{tabular}[c]{@{}l@{}}Support Vector \\ Machine (SVM)\end{tabular} \\ \hline
\begin{tabular}[c]{@{}l@{}}Probst et al. {\cite{ma3}}\end{tabular} &2015&
  - &
  - &
  \begin{tabular}[c]{@{}l@{}}SIFT, LBP \\ and Color\end{tabular} &
  Linear SVM \\ \hline
\begin{tabular}[c]{@{}l@{}}Meyers et al. {\cite{ma4}}\end{tabular} &2015&
  Im2Calories &
  Yes &
  GoogleNet CNN &
  \begin{tabular}[c]{@{}l@{}}GoogleNet CNN \\ model\end{tabular} \\ \hline
\begin{tabular}[c]{@{}l@{}}Ravi et al. {\cite{ma5}}\end{tabular} &2015&
  FoodCam &
  No &
  \begin{tabular}[c]{@{}l@{}}HoG, LBP and \\ RGB Color   \\ Features\end{tabular} &
  Linear SVM \\ \hline
\begin{tabular}[c]{@{}l@{}}Waltner et al. {\cite{ma6}}\end{tabular} &2017&
  - &
  Yes &
  \begin{tabular}[c]{@{}l@{}}RGB, HSV and \\ LAB Color \\ values\end{tabular} &
  \begin{tabular}[c]{@{}l@{}}Random Forest \\ Classifier\end{tabular} \\ \hline
\begin{tabular}[c]{@{}l@{}}Mezgec and Seljak {\cite{ma7}}\end{tabular} &2017&
  - &
  - &
  NutriNet &
  NutriNet \\ \hline
\begin{tabular}[c]{@{}l@{}}Pouladzadeh et al. {\cite{ma8}}\end{tabular} &2017&
  - &
  Yes &
  CNN &
  \begin{tabular}[c]{@{}l@{}}Caffe \\ Framework\end{tabular} \\ \hline
\begin{tabular}[c]{@{}l@{}}Waltner et al.   {\cite{ma9}}\end{tabular} & 2017&
  - &
  Yes &
  CNN &
  CNN \\ \hline
\begin{tabular}[c]{@{}l@{}}Ming et al. {\cite{ma10}}\end{tabular} &2018&
  DietLens &
  - &
  ResNet-50 CNN &
  ResNet-50 CNN \\ \hline
\begin{tabular}[c]{@{}l@{}}Jiang et al.  {\cite{ma11}}\end{tabular} &2018&
  - &
  Yes &
  \begin{tabular}[c]{@{}l@{}}Colors, Lines, \\ Points,SIFTand \\ Texture Features\end{tabular} &
  \begin{tabular}[c]{@{}l@{}}Reverse Image Search   \\ (RIS) and Text Mining\end{tabular} \\ \hline
\begin{tabular}[c]{@{}l@{}}Jianing Sun et al. {\cite{ma12}}\end{tabular} &2019&
  Food Tracker &
  Yes &
  DCNN &
  DCNN \\ \hline
\begin{tabular}[c]{@{}l@{}}G. A. Tahir \\ and C.K. Loo \cite{d42} \end{tabular} & 2020&
  MyDietCam &
  Yes &
  \begin{tabular}[c]{@{}l@{}}ResNet-50, \\ DenseNet201   \\ and Inception \\ ResNet-V2\end{tabular} &
  \begin{tabular}[c]{@{}l@{}}Adaptive Reduced \\ Class Incremental \\ Kernel Extreme \\ Learning Machine \\ (ARCIKELM)\end{tabular} \\ \hline
\end{tabular}
}
\end{table*}

\subsection{Harnessing Vision-Based Method to Measure Nutrient Intake During COVID-19}
As the covid-19 is a leading global challenge across the world, maintaining good nutritional status is mandatory for keeping good health to fight against the virus. Automatic vision-based methods for volume estimation and food image recognition in these nutrition tracking apps can assist patients to objectively measure nutrient intake of vital vitamins required for boosting the immune system.
\subsection{Life's Simple 7}
Life’s Simple 7 health score is recently introduced based on modifiable health factors which contribute to heart health. Physical activity, non-smoking, healthy diet and body mass index are four modifiable health behaviours in it. The other three modifiable factors are biological. It includes blood pressure, fasting glucose and cholesterol details. Besides cardiovascular health, Life’s Simple 7 also impacts other health conditions such as Venous Thromboembolism, cognitive health, Atherosclerosis etc. As dietary intake plays a vital role in computing life's simple 7,  manually these factors and then calculating a life simple 7 score is a very tedious process. It makes it very difficult for both middle-aged patients and elderly patients to keep track of their health. So vision-based methods can play an important role in automating the diet score. However, there are no current studies that have explored this research direction.
\subsection{Enforcing Eating Ban on Public Places During COVID-19 Pandemic or Other Restricted Places}
Vision-based food recognition can automate the enforcement of an eating ban at public places by automatically detecting foods from CCTV AND wearable cameras to curb the spread of the virus. Similarly, vision-based food recognition coupled with CCTV or wearable cameras and smart apps automate the enforcement of eating ban at workplaces, laboratories etc. 
\subsection{Monitoring Malnutrition in Low Income Countries}
Coupling vision-based methods with wearable cameras can automatically detect foods from egocentric images with reasonable accuracy while reducing the burden of processing big data and addressing the user's privacy concerns. Egocentric images acquired from these cameras are important to study diet and lifestyle, especially in low-income countries with a high malnutrition rate.  For example, Jia et al. \cite{app2} focused on gathering image data from wearable cameras and discriminating between food/non-food classes based on their tag from the CNN to study human diet. Similarly, Chen et al. \cite{app1} studied malnutrition in low and middle-income countries by using the wearable device e-button.
\subsection{Food Image Analysis From Social Media}
We are in the era of social media, and food is a basic necessity of life, a great deal of content on social media platforms are related to food. User's of these platforms frequently share new recipes, new methods of cooking, food pictures after restaurant check-in. Researchers have exploited this data on social media platforms for analyzing dietary intake. For example, Mejova et al. \cite{app3} studied food images from foursquare and Instagram to analyze the food consumption pattern in the USA. Similarly, food images on social media platforms are of different cultures. These images can be crawled and then combined together to prepare a large food database.
\subsection{Food Quality Assessment}
Evaluating fruits quality, freshness at the marketplace and the user end is of increasing interest as opposed to accessing quality at the time of manufacturing. Efforts to date have focused on accessing the quality of foods using vision-based methods. For example, Ismail et al. has contributed an Apple-NDDA dataset \cite{app4} which consists of defective and non-defective apple images for food quality assessment 

\section{Statistical Analysis}
\begin{figure}[!htbp]
\centering
\includegraphics[width=3.5in]{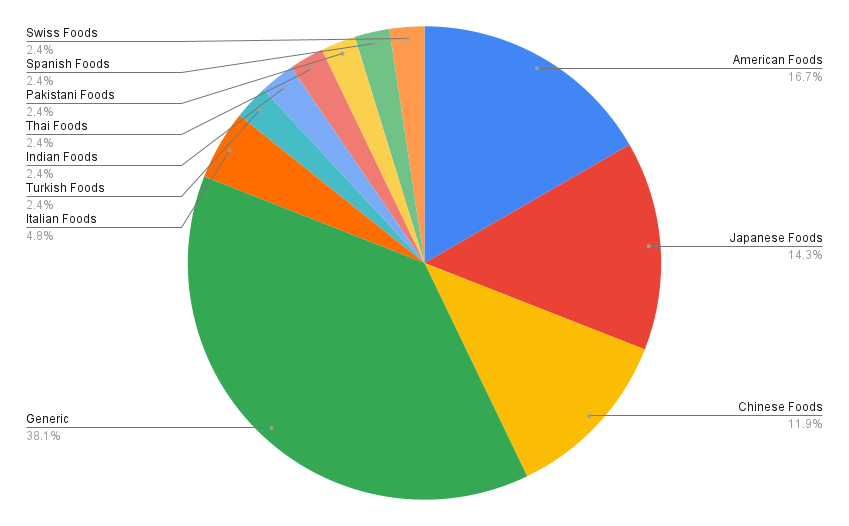}
\caption{Volume estimation methods using single image vs multiple images}
\label{fig_foodcountries}
\end{figure}

\begin{figure}[!htbp]
\centering
\includegraphics[width=3.5in]{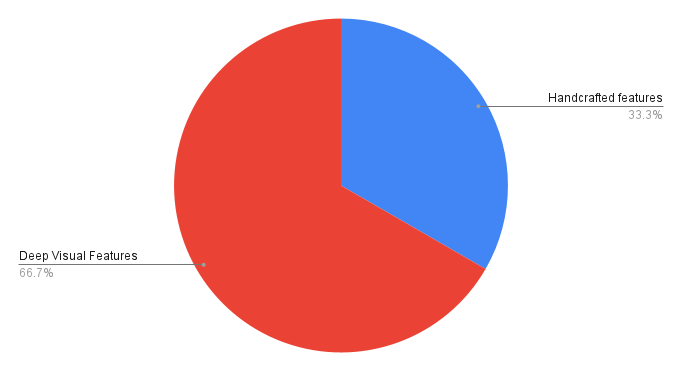}
\caption{Percentage of studies summarized according to the type of feature extraction methods.}
\label{fig_pie_featurecategory}
\end{figure}

\begin{figure}[!htbp]
\centering
\includegraphics[width=3.5in]{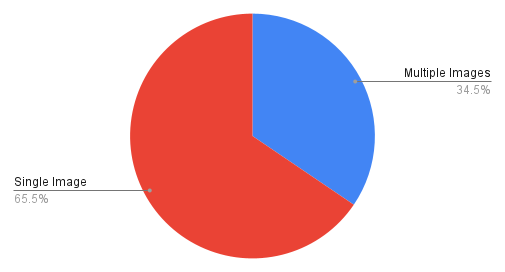}
\caption{Volume estimation methods using single image vs multiple images}
\label{fig_single_multiple}
\end{figure}

We exhibit statistical analysis of our study based on the articles and conference proceedings gathered to write this survey paper. We surveyed research studies up to 2020 from various reputed sources IEEE, Elsevier, ACM and web of sciences.  Figure \ref{fig_foodcountries} shows the pie chart of the distribution of surveyed food databases according to the country to which food dishes belongs. In it, Generic databases are those which contain food dishes of multiple countries. We summarized the surveyed studies in two main categories, studies using handcrafted features and studies using visual features representations from the convolutional neural networks (CNN), as shown in Figure \ref{fig_pie_featurecategory}. As discussed in section 1.1, volume estimation methods require a single view or multiple images from different viewpoints. We presented a pie chart as shown in Figure \ref{fig_single_multiple} that describes the percentage of studies we surveyed according to the number of image viewpoints required to estimate food volume. For ingredient detection, all included studies are using CNN due to recent interest in this extension. Similarly, for studies that have implemented mobile applications, the piechart in figure 8 shows that 46.2\% of applications are implementing CNN for food recognition while 45.2\% of mobile applications from surveyed studies are implementing traditional methods for feature extraction.

\begin{figure}[!htbp]
\centering
\includegraphics[width=3.5in]{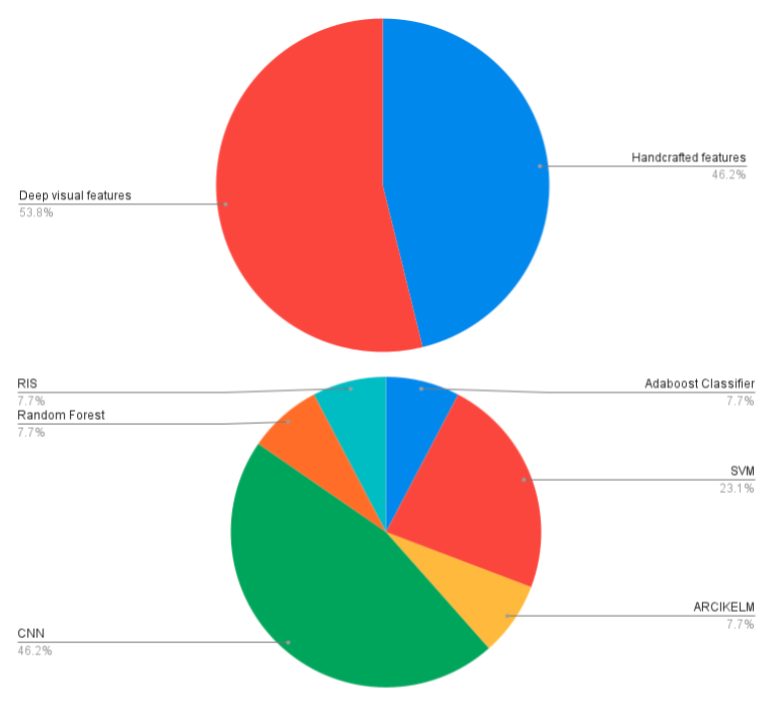}
\caption{Percentage of studies summarized according to the type of methods employed for feature extraction from food images and category of classifier used for food image analysis in a mobile application.}
\label{fig_piechart_mobileapps}
\end{figure}

\section{Open Issues}
We highlighted open issues based on the survey papers and the author's first-hand experience with existing methodologies.
\subsection{Unsupervised Learning From Unlabelled Dataset}
Preparing a large comprehensive annotated data is still a challenge, as manually annotating a dataset is a difficult task with many challenges. Due to the large variety of food dishes, different styles of preparation etc., it is difficult for an expert dietician to correctly label all the foods especially, in the preparation of a multi-culture food database. Similarly, it involves high costs and a large number of working hours to prepare such a dataset.  Recent advancements in contrastive learning have opened a new research paradigm of unsupervised learning. Methods based on contrastive learning such as SimCLR \cite{oissue1} and SwAV \cite{oissue2} does not require labelled datasets and seems to be interesting potential areas of research that future works in food recognition should exploit.
\subsection{Continual Learning}
Food datasets are open-ended, and there is no cap on the number of dishes. So the network must adapt to continuously evolving datasets. All of these properties of food datasets have made them a strong use case for continual learning methods. One of the principal challenges in continuous learning methods is catastrophic forgetting. Catastrophic forgetting refers to completely or abruptly forget previously learned information while learning new classes. Many neural networks are susceptible to forgetting during continual learning. It is a prime hindrance to achieve the objective of continuously evolving networks similar to humans.  Hence, researchers should also study catastrophic forgetting in context with food databases.
\subsection{Explainability}
Although, there have been numerous including, activation methods, SHAP values \cite{oissue4} and distillation methods. There is still a research gap in the context of food recognition. As food recognition has many domain-specific challenges such as intraclass variations, non-rigid structure, visualization of the reasoning behind model predictions is vital to trust its decisions. Recently unsupervised clustering methods \cite{oissue5} are exploited to explain model predictions by distilling knowledge into surrogate models. It provides similar images to test images for explaining prediction results. Explaining prediction results by showing images similar to test images seems more friendly as users do not need any specific domain knowledge to understand these results.
\section{Discussion}
Our research provides deep insight into computer vision-based approaches for dietary assessment. It focuses on both traditional and deep learning methodologies for feature extraction and classification methods used for food image recognition and single and multiview methods for volume estimation. Similarly, this survey also explores and compares current food image datasets in detail, as vision-based techniques are highly dependent on a comprehensive collection of food images. In contrast to previous research work, Mohammad A. Sobhi et al. \cite{dis1}, Min, Weiqing, et al.\cite{dis2}. Our survey scrutinizes traditional and current deep visual approaches for feature extraction and classification to enhance clarity in terms of their performance and feasibility. 
Unlike existing surveys, our survey emphasizes existing solutions developed for food ingredient recognition through multi-label learning. We also reviewed existing computer-based food volume estimation methods in detail as they have reduced dietitians and experts intervention and can accurately determine the portion size of the food in contrast to the self-estimation. 
Finally, our research study also explores real-world applications using the prior methodologies for dietary assessment purposes.  
\subsection{Findings}
Our findings indicate that the ultimate performance of traditional and deep visual techniques depends on the type of dataset used. Therefore, it has been observed out of 38 datasets (as shown in Table \ref{datasets}) explored in this survey, 3 most commonly used datasets were UECFOOD-256 \cite{hf11}, UECFOOD-100 \cite{dataset2} and Food-101 \cite{d17}. UECFOOD-256 (25088 images and 256 classes) and UECFOOD-100 (14361 images and 100 classes of food) are Japanese food datasets consisting of Japanese Foods images captured by users, whereas Food-101(101000 images and 101 classes) is an American fast food dataset containing images crawled from several websites. However, these widely used datasets are region-specific.  Therefore, there is an immense need for generic food datasets for excluding regional bias from experimental results. 
Besides this, it is also evident from this survey that deep visual techniques have replaced traditional machine learning methodologies for food image recognition. As per our survey, systems proposed after the year 2015 are mainly using deep learning technologies for food classification purposes. It is due to their phenomenal classification performance. Speaking of classification performance of deep visual techniques, for food-non food classification McAllister et al. 2018 \cite{d32} (99.4\%) and Pouladzadeh et al., 2016 \cite{d25} (99\%) achieved the highest top 1 classification accuracy. Pouladzadeh et al., 2016 \cite{d25} used DCNN and Graph cut on their proposed dataset, whereas McAllister et al. 2018 \cite{d32} used CNN, ANN, SVM, and random forest on the food 5k dataset. Table \ref{classification Accuracies of deep models} further compares classification accuracies of proposed deep visual models.
Recent advancements and exceptional performance of food image classification methods have now led researchers to explore food images from a much deeper perspective in terms of retrieval and classification of food ingredients from food images. Therefore, we have also explored several proposed solutions for food ingredient recognition and classification. According to our survey, the system proposed by Chen et al.,2016 \cite{hclassification1} has achieved the highest F1 score i.e. 95.88\% macro-F1 and 82.06\% micro-F1 using the Arch-D method on the UECFOOD-100 dataset (as shown in Table \ref{multilabel}). 
Similarly, automatic food volume estimation methods have reduced dietitians and experts intervention and can accurately determine the portion size of the food in contrast to the self-estimation for food volume estimation. Single view methods involve capturing a single image while multi-views require multiple images to determine accurate food volumes. Results in Table \ref{multiview} depicts that multi-view methods are mostly better than single-view methods.

Finally, food category recognition, ingredient classification, and volume estimation techniques helped provide an automatic dietary assessment with reduced human intervention in mHealth apps. For this purpose, we have also surveyed several mobile applications which employ deep learning methods for dietary assessment. 

\subsection{Limitations and Future Research Challenges}
Despite enhanced performance and classification accuracy, food image recognition and volume estimation through vision-based approaches may continue to present interesting future research challenges. This is because the performance of the methodologies used for food image identification is highly dependent on the source of images in a particular food dataset. Although the growing number of food categories are being incorporated in food image datasets like UECFOOD-256 \cite{hf11}, Food 85 \cite{hf1}, and Food201-segmented \cite{ma4}, there is still an immense need for generalized, comprehensive datasets for better performance evaluation and benchmarking. Moreover, we observed that datasets with a large number of food images significantly positively impact classification accuracy. However, keeping these large image datasets updated is another challenge, especially since different types of foods are being prepared every day. 
\par In addition to this, progressive learning during the classification phase is vital for food image datasets due to the continuous arrival of new concepts and domain variation within existing concepts. Similarly, developing frameworks interpretable by highlighting the contribution of the area of interest will improve the overall human trust level on a solution in a real-world environment. 
\par Following food recognition, food volume estimation is a particularly complex and challenging assignment since food items have large variations in shape, texture and appearances. Since our article categorized food portion estimation methods into a single view and multi-view methods. Multi-view methods are more accurate, however, most of these methods also require calibration objects each time and images from different viewpoints that make the usability of these solutions tedious for elderly users.

\par Finally, there is a need to design and develop solutions that can respond to situations ethically. In our context, it refers to the removal of any biases concerning region-specific food preferences. It will help to ensure transparency in existing models.

\section{Conclusion}
In this work, we explored a broad spectrum of vision-based methods that are specifically tailored for food image recognition and volume estimation. In practice, the food recognition process incorporates four tasks: acquiring food images from the corresponding food datasets, feature extraction using handcrafted or deep visual, selection of relevant extracted features, and finally, appropriate selection of classification technique using either traditional machine learning approach or deep learning models followed by food ingredient classification to provide better insight of nutrient information. 
\par Despite, impeccable performance exhibited by state-of-the-art approaches, there exists several limitations and challenges. There is an immense need for comprehensive datasets for benchmarking and performance evaluation of these models, as incorporating large food image datasets improves the overall performance.  Consequently, when dealing with open-ended and dynamic food datasets, the classifier must be capable of open-ended continuous learning. However, existing methods have several bottlenecks, which undermine the food recognition ability when it comes to open-ended learning as proposed methods are prone to catastrophic forgetting. They tend to forget previous knowledge extracted from images while learning new information. Such methods work great only for fixed food image datasets. Moreover, proposed techniques for food ingredient classification are still struggling with performance issues when applied to prepared and mixed food items. 
\par Similarly, automatic food portion estimation methods are categorized into two major categories single view image methods and multi-view image methods. As discussed earlier, most of the multi-view image methods are more accurate as compared to the single view methods but multi-view image methods require complex processing and images from different angles resulting in a reduced user retention rate. Furthermore, most of the single and multi-view methods require calibration objects each time, which makes the usability of these solutions tedious for elderly patients. 
\par Therefore, there is substantial room for innovative health care and dietary assessment applications that can integrate wearable devices with a smartphone to revolutionize this research area. Moreover, dietary assessment systems should address these challenges to provide better insights into effective health maintenance and chronic disease prevention.
\section*{Author Contributions}
Ghalib Ahmed Tahir was responsible for the literature search and writing the article and approved the final version as submitted.  Loo Chu Kiong contributed to the study design, reviewed the study for intellectual content, and confirmed the final version as submitted.
\section*{Conflict of interest}
The authors wish to confirm that there are no conflicts of interest.
\section*{Acknowledgement}
This research was supported by the UM Partnership Grant: Project No: RK012-2019 from University of Malaya, IIRG Grant (IIRG002C-19HWB) from University of Malaya, International Collaboration Fund for project Developmental Cognitive Robot with Continual Lifelong Learning (IF0318M1006) from MESTECC, Malaysia and  ONRG grant (Project No.: ONRG-NICOP- N62909-18-1-2086) / IF017-2018 from Office of Naval and Research Global, UK.


\end{paracol}
\reftitle{References}

\end{document}